\documentclass[acmlarge, screen]{acmart}

\usepackage{algorithmic}
\usepackage{algorithm}
\usepackage{textcomp}
\usepackage{stfloats}
\usepackage{url}
\usepackage{verbatim}
\usepackage{multicol}
\usepackage{multirow}
\usepackage{xcolor} 
\usepackage{colortbl}

\usepackage{amsmath}  
\usepackage{bm}       
\usepackage{tabularx}
\usepackage{makecell} 
\usepackage{rotating} 
\usepackage{soul}     
\usepackage{longtable}
\usepackage{mathtools}
\usepackage{float}
\usepackage{array}    
\usepackage{booktabs} 
\usepackage{geometry}
\usepackage{graphicx}
\RequirePackage{fontawesome}
\usepackage{pdflscape} 

\usepackage{hyperref}

\AtBeginDocument{%
  \providecommand\BibTeX{{%
    \normalfont B\kern-0.5em{\scshape i\kern-0.25em b}\kern-0.8em\TeX}}}

\setcopyright{acmcopyright}
\copyrightyear{20XX}
\acmYear{202X}
\acmDOI{XXXXXXX.XXXXXXX}

\acmJournal{CSUR}
\acmVolume{00}
\acmNumber{0}
\acmArticle{000}
\acmMonth{0}

\begin{document}

\title[Mamba in Vision: A Comprehensive Survey of Techniques and Applications]{Mamba in Vision: A Comprehensive Survey of Techniques and Applications}

\author{Md Maklachur Rahman}
\orcid{0000-0002-1984-767X}
\email{maklachur@tamu.edu}
\affiliation{%
  \institution{Texas A\&M University}
  \city{College Station, Texas}
  \country{USA}
  \postcode{77843}
  }
\author{Abdullah Aman Tutul}
\orcid{0000-0002-7931-2829}
\email{abdullahaman633@tamu.edu}
\affiliation{%
  \institution{Texas A\&M University}
  \city{College Station, Texas}
  \country{USA}
  \postcode{77843}
  }
\author{Ankur Nath}
\orcid{0009-0007-9145-9894}
\email{anath@tamu.edu}
\affiliation{%
  \institution{Texas A\&M University}
  \city{College Station, Texas}
  \country{USA}
  \postcode{77843}
  }
\author{Lamyanba Laishram}
\orcid{0000-0002-0324-214X}
\email{yanbalaishram@gmail.com}
\affiliation{%
  \institution{Kyungpook National University}
  \city{Daegu}
  \country{South Korea}
  \postcode{41566}
}
\author{Soon Ki Jung}
\orcid{0000-0003-0239-6785}
\email{skjung@knu.ac.kr}
\affiliation{%
  \institution{Kyungpook National University}
  \city{Daegu}
  \country{South Korea}
  \postcode{41566}
}
\author{Tracy Hammond}\authornote{Corresponding Author}
\orcid{0000-0001-7272-0507}
\email{hammond@tamu.edu}
\affiliation{%
  \institution{Texas A\&M University}
  \city{College Station, Texas}
  \country{USA}
  \postcode{77843}
  }

\renewcommand{\shortauthors}{Rahman et al.}

\begin{abstract}
Mamba is emerging as a novel approach to overcome the challenges faced by Convolutional Neural Networks (CNNs) and Vision Transformers (ViTs) in computer vision. While CNNs excel at extracting local features, they often struggle to capture long-range dependencies without complex architectural modifications. In contrast, ViTs effectively model global relationships but suffer from high computational costs due to the quadratic complexity of their self-attention mechanisms. Mamba addresses these limitations by leveraging Selective Structured State Space Models to effectively capture long-range dependencies with linear computational complexity. This survey analyzes the unique contributions, computational benefits, and applications of Mamba models while also identifying challenges and potential future research directions. We provide a foundational resource for advancing the understanding and growth of Mamba models in computer vision. An overview of this work is available at \textcolor{blue}{\textit{https://github.com/maklachur/Mamba-in-Computer-Vision}}.
\end{abstract}

\ccsdesc[500]{General and reference~Surveys and overviews}
\ccsdesc[500]{Computing methodologies~Learning paradigms}
\ccsdesc[500]{Computing methodologies~Deep Learning approaches}
\ccsdesc[500]{Computing methodologies~Computer Vision algorithms}

\keywords{state space model, mamba, mamba models, vision mamba, computer vision}


\maketitle
\section{Introduction}
\label{sec:introduction}
The evolution of deep learning has significantly advanced computer vision, with  \textbf{Convolutional Neural Networks (CNNs)} \cite{krizhevsky2012imagenet} playing a key role. CNNs revolutionized the field by enabling machines to learn complex patterns directly from pixel data through convolutional layers that capture features at multiple scales and build spatial hierarchies. Despite their success, CNNs encounter inherent challenges in capturing long-range dependencies due to their localized receptive fields. Addressing these challenges often requires deeper and more complex architectures, which increase computational cost and reduce efficiency \cite{rahman2020efficient, he2016deep, szegedy2016rethinking}.

To improve sequence modeling and global context understanding,  \textbf{Recurrent Neural Networks (RNNs)} \cite{schmidt2019recurrent} were initially developed, followed by the introduction of Transformers \cite{vaswani2017attention}, which brought significant breakthroughs in deep learning. RNNs, especially those with  \textbf{Long Short-Term Memory (LSTM)} units \cite{hochreiter1997long}, improved the ability to capture temporal dependencies in sequential data. However, their sequential nature limits parallel processing, which slows down speed and reduces scalability \cite{hochreiter1997long, cho2014learning}.
With their self-attention mechanisms, Transformers overcame this limitation by allowing models to dynamically prioritize different parts of the input data \cite{vaswani2017attention}. The \textbf{Vision Transformer (ViTs)} were developed for images and treated them as sequences of patches, capturing global dependencies more effectively than CNNs \cite{dosovitskiy2020image, Rahman_2024, khan2022transformers}. However, despite their strong performance in various computer vision tasks, ViTs face computational efficiency challenges due to the quadratic complexity of their self-attention mechanism, especially in high-resolution and real-time applications \cite{Rahman_2024}.

Hybrid models have emerged to address the limitations of traditional architectures by integrating the strengths of CNNs, RNNs, and Transformers in computer vision tasks. For instance, \textbf{Convolutional LSTMs (ConvLSTMs)} \cite{shi2015convolutional} enhance the model’s ability to capture spatial-temporal relationships by integrating convolutional operations within LSTM units \cite{shi2015convolutional}. Similarly, MobileViT merges the local feature extraction of CNNs with the global context modeling of Transformers \cite{mehta2021mobilevit}. Hybrid architectures aim to balance high performance and computational efficiency but add complexity due to component optimization requirements.

Recently, \textbf{State Space Models (SSMs)} have gained attention as a promising alternative, especially for handling sequential data where efficiently managing long-range dependencies is crucial \cite{gu2020hippo, gu2021combining, gu2021efficiently}. The \textbf{Structured State Space Sequence (S4)} model is a notable development in this domain, which leverages state space representations to achieve linear computational complexity. As a result, long sequences can be processed efficiently without compromising accuracy \cite{gu2021efficiently}. The S4 model achieves this by integrating both recurrent and convolutional operations, which helps reduce the computational demands typically associated with sequence modeling.

Building on the foundational principles of SSMs, the \textbf{Mamba} model \cite{gu2023mamba} represents a significant leap in sequence modeling. Mamba integrates state space theory with advanced deep learning techniques, employing selective state representations that dynamically adjust based on the input data. This selective state mechanism dynamically filters out less important information to focus on the most relevant parts of the input sequence, thereby reducing computational overhead and enhancing efficiency \cite{gu2023mamba}. The Mamba architecture utilizes a hardware-aware, scan-based algorithm optimized for GPUs, avoiding the inefficiencies of traditional convolution-based SSMs. This leads to faster training and inference, enabling more efficient handling of visual data and a transformative approach to computer vision \cite{zhu2024vision}.

\begin{figure*}[]
  \centering
  \includegraphics[scale=0.22, keepaspectratio]{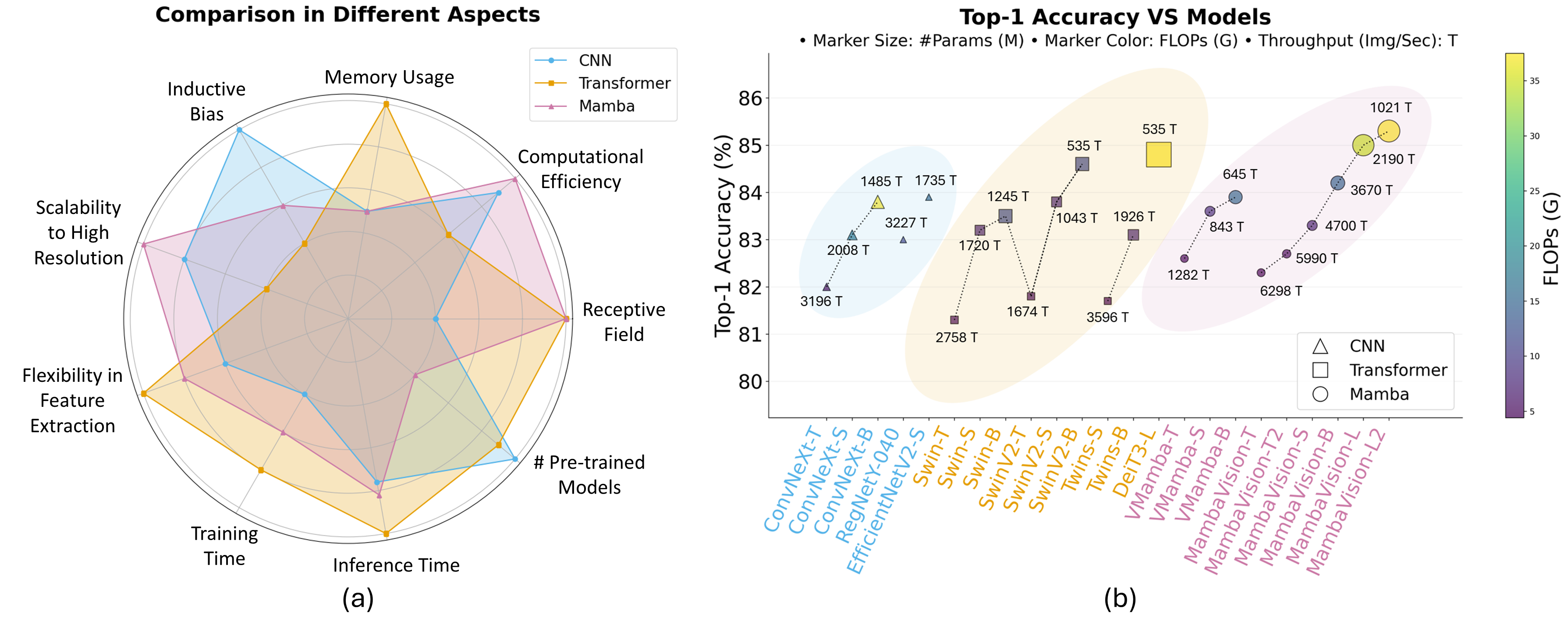}
  \caption{(a) Comparative trade-offs of CNN, Transformer, and Mamba frameworks. (b) Comparison of Top-1 accuracy for CNN, Transformer, and Mamba models on ImageNet-1K \cite{hatamizadeh2024mambavision}.}
  \label{fig:trade_off_comparison} \vspace{-3.4mm}
\end{figure*} 

\begin{table}[]
\centering
\caption{Overall Comparison of Survey Papers on Mamba Models}
\label{tab:survey_comparison}
\resizebox{\columnwidth}{!}{%
\begin{tabular}{@{}cccccccc@{}}
\toprule
\multirow{2}{*}{\textbf{Survey Papers}}       & \multirow{2}{*}{\textbf{Domain}} & \multirow{2}{*}{\textbf{Taxonomy}} & \textbf{Unique Aspects}  & \textbf{Distinctive Features} & \textbf{Fundamental Distinctions among} & \textbf{Experimental}   & \textbf{Challenges \& Future} \\
                                              &                                  &                                    & \textbf{of Mamba Models} & \textbf{in Scanning}          & \textbf{CNN, Transformer, and Mamba}   & \textbf{Analysis}       & \textbf{Prospects Analysis}   \\ \midrule
\cite{wang2024state} [Apr 2024, arXiv]   & Generic                          & \checkmark                         &                          &                               &                                        & \checkmark              & \checkmark                   \\ 
 \midrule
\cite{patro2024mamba} [Apr 2024, arXiv]  & Generic                          & \checkmark                         &                          &                               & \checkmark                             & \checkmark              &                               \\ 
 \midrule
\cite{qu2024survey} [Aug 2024, arXiv]  & Generic                          & \checkmark                         &                          & \checkmark                    & \checkmark                             &                        & \checkmark                   \\ 
 \midrule
\cite{heidari2024computation} [June 2024, arXiv] & Medical                     & \checkmark                         & \checkmark               &                               &                                        & \checkmark              & \checkmark                   \\ 
                           \midrule
\cite{mamba_scanning} [Apr 2024, arXiv]   & Remote Sensing                   &                                    &                          &                               &                                        & \checkmark              &                               \\ 
 \midrule
\cite{xu2024survey} [Apr 2024, arXiv]      & Vision                           & \checkmark                         &                          &                               &                                        & \checkmark              & \checkmark                   \\ 
 \midrule
\cite{liu2024vision} [Apr 2024, arXiv]    & Vision                           & \checkmark                         &                          &                               &                                        & \checkmark              & \checkmark                   \\ 
 \midrule
\cite{zhang2024survey} [June 2024, MDPI] & Vision                           & \checkmark                         &                          &                               &                                        &                        &                               \\ 
 \midrule
Ours & Vision                           & \checkmark                         & \checkmark                     &    \checkmark                     &    \checkmark                              &    \checkmark          &    \checkmark               \\ 
 
\bottomrule
\end{tabular}%
}
\end{table}
\setlength{\textfloatsep}{10pt}

Mamba models are particularly advantageous for tasks such as video processing \cite{chaudhuri2024simba}, for long temporal sequences processing; remote sensing \cite{chen2024changemamba}, for large spatial datasets; and medical imaging \cite{wang2024mamba}, for efficient and precise high-resolution data processing. CNNs and Transformers face scalability issues due to high computational demands, while Mamba models overcome this by offering linear scalability with sequence length, making them ideal for real-time and large-scale applications. Their integration of state space principles with selective attention mechanisms offers a robust approach to handling complex visual tasks, enabling more efficient and scalable computer vision solutions. Figure \ref{fig:trade_off_comparison}(a) qualitatively compares CNN, Transformer, and Mamba frameworks, while Figure \ref{fig:trade_off_comparison}(b) provides a quantitative comparison on the ImageNet-1K \cite{ILSVRC15} dataset based on various metrics.
 
While recent survey papers have explored various aspects of Mamba models—such as SSMs \cite{wang2024state}, applications in computer vision \cite{zhang2024survey, xu2024survey, liu2024vision}, and medical image analysis \cite{heidari2024computation}—our paper provides a unique perspective by distinguishing itself in areas like model taxonomy, scanning methods, application domains, comparative analysis with CNNs and Transformers, and future directions, as outlined in Table \ref{tab:survey_comparison}.

The main contributions of our work are summarized as follows:
\begin{itemize}
    \item We provide a comprehensive overview of Mamba models in computer vision, highlighting their distinctive features and comparative analysis with CNNs and Transformers.
    
    \item We present a novel taxonomy categorizing Mamba models by application areas in computer vision, guiding researchers in selecting appropriate models for their specific needs.

    \item We demonstrate the strengths and weaknesses of the Mamba model's core component, the scanning methods, and their specific use cases. 

    \item  Finally, we outline key challenges in Mamba models and propose future research directions to further improve their application in computer vision.

\end{itemize} 

The paper is structured as follows: Section \ref{sec:taxonomy} outlines the taxonomy of Mamba models, followed by an overview of the Mamba-based vision pipeline and scanning techniques in Section \ref{sec:mamba_models}. Applications across different domains are discussed in Section \ref{sec:mamba_in_cv}, with a comparative analysis of Mamba, Transformer, and CNN models presented in Section \ref{sec:comparative_analysisV}. Section \ref{sec:limitations_mamba_V} explores potential limitations and future research directions, while Section \ref{sec:conclusion_V} concludes the paper by summarizing key findings.
\section{Taxonomy of Mamba Models} \label{sec:taxonomy}

\begin{figure*}[t]
  \centering
  \includegraphics[width=\linewidth]{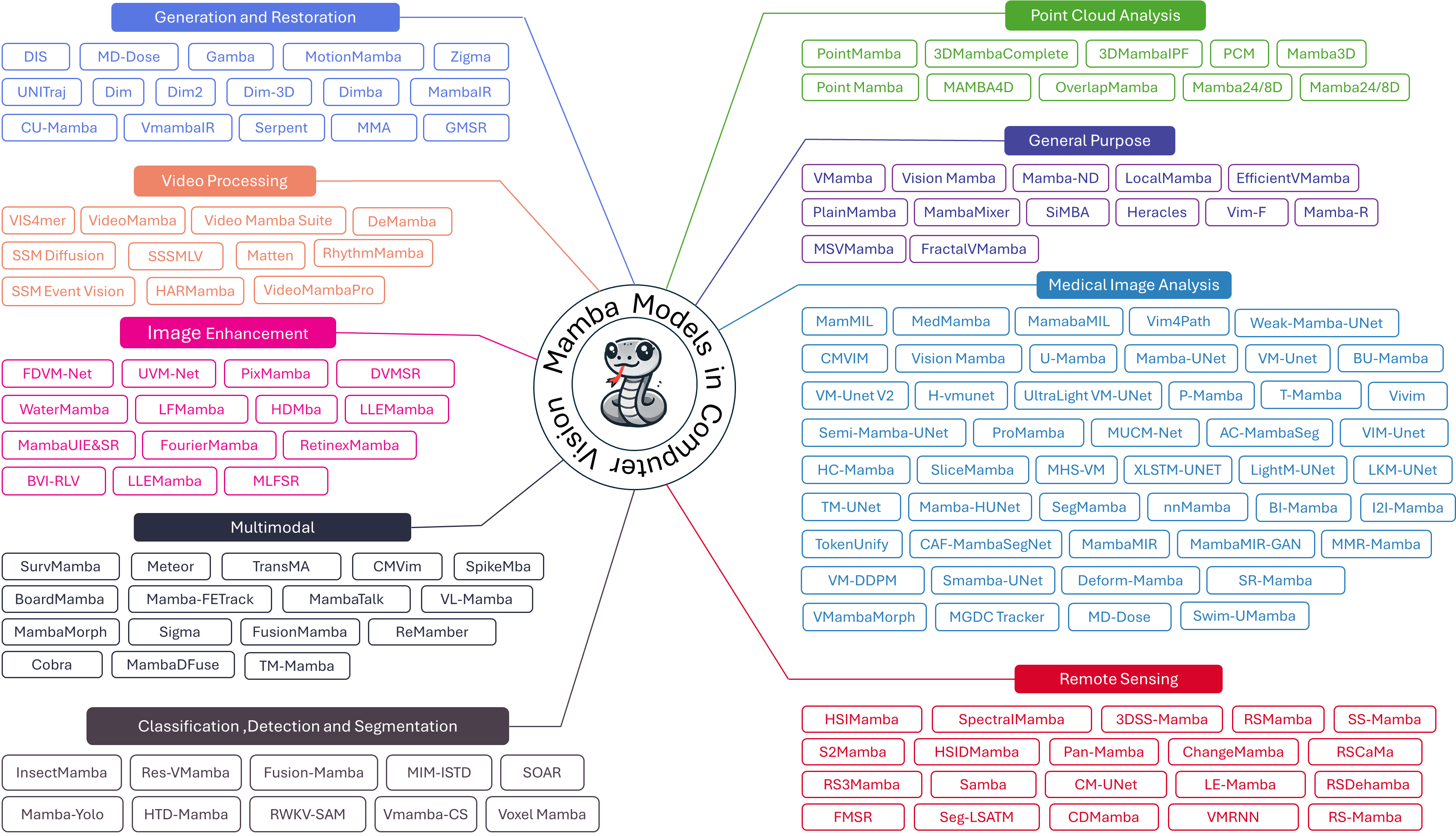} 
  \caption{Overall taxonomy of Mamba models in computer vision tasks, categorized by their application areas. This taxonomy includes models from the baseline Mamba model \cite{gu2023mamba} up to those published by July 15, 2024.}
  \label{fig:taxonomy} 
\end{figure*}

The adaptation of Mamba for visual tasks began in early 2024, with models like VMamba \cite{liu2024vmamba} and Vision Mamba (Vim) \cite{zhu2024vision}. These initial models have pushed the boundaries of visual processing, providing efficient solutions to complex challenges. For the convenience of future researchers, we have developed a comprehensive taxonomy, as shown in Figure \ref{fig:taxonomy}.  This categorization highlights the diverse applications of Mamba-based models across nine categories, with significant contributions to medical image analysis. 

\setlength{\textfloatsep}{10pt}

\section{Overview of Mamba Models}
\label{sec:mamba_models}
 \begin{figure*}[t!]
    \centering
    \includegraphics[scale=0.3]{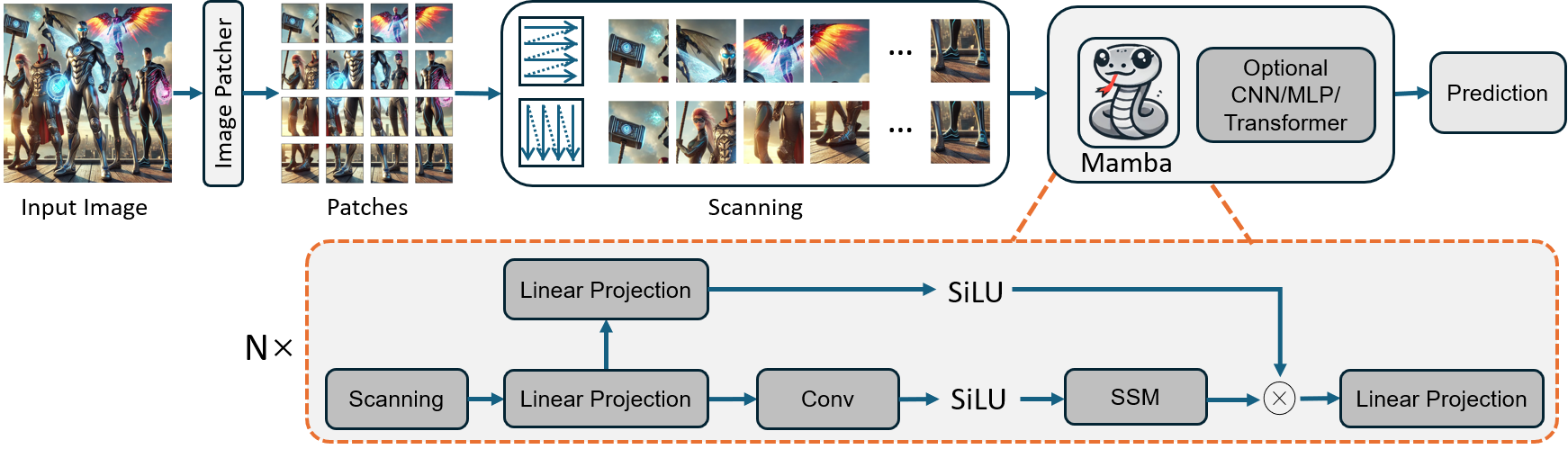}
    \caption{Overall pipeline of a Mamba-based vision model.}
    \label{fig:mamba_pipeline_fig3}
 \end{figure*}

In this section, we provide the basic pipeline for Mamba-based vision models in Figure \ref{fig:mamba_pipeline_fig3}. This pipeline starts with patching the input image, followed by a series of scanning operations designed to extract multi-scale features. The patches are then passed through a Mamba block, which typically consists of linear projections, convolutional layers, SiLU activation, and SSM operations to extract features optimally. Depending on the specific task, many models then integrate CNN and transformer blocks to enhance performance. Now, we explore the inner workings of the Mamba block in the following section.

\subsection{Fundamentals of Mamba Models}
\label{sec:mamba_fundamentals}
This section provides an overview of the structure and technical details of Mamba models.

\subsubsection{State Space Model}
SSMs are a fundamental class of models in deep learning for sequence data. These models are designed to map a one-dimensional input sequence, denoted as $x(t)$ and existing in a real vector space $\mathbb{R}^L$, to an output sequence $y(t)$ in the same space, via an intermediate latent state $h(t)$ residing in $\mathbb{R}^N$. The dynamics of these models are controlled by a set of linear transformations \cite{gu2023mamba}, as described by the equations:

\begin{equation}
\begin{gathered}
h'(t) = \mathbf{A}h(t) + \mathbf{B}x(t), \quad 
y(t) = \mathbf{C}h(t).
\end{gathered}
\end{equation}

Here, $\mathbf{A}$, $\mathbf{B}$, and $\mathbf{C}$ are system matrices of appropriate dimensions that dictate the state transitions, input, and output mapping, respectively. These matrices are defined as $\mathbf{A} \in \mathbb{R}^{N \times N}$, $\mathbf{B} \in \mathbb{R}^{N \times 1}$, and $\mathbf{C} \in \mathbb{R}^{N \times 1}$.

For practical applications, the continuous-time model needs to be converted into a discrete-time model for implementation in digital systems. This discretization is typically achieved using a zero-order hold assumption, where the continuous-time system parameters $\mathbf{A}$ and $\mathbf{B}$ are transformed into their discrete counterparts over a sampling timescale $\Delta$, which is a positive real number. The discretized system is represented as:

\begin{equation}
\bm{\overline{A}} = e^{\bm{\Delta} \bm{A}}, \quad
\bm{\overline{B}} = (\bm{\Delta} \bm{A})^{-1} (e^{\bm{\Delta} \bm{A}} - \bm{I}) \cdot \bm{\Delta} \bm{B}.
\end{equation}

\noindent The resulting discrete model equations are:
\begin{equation}
\begin{gathered}
h_t = \bm{\overline{A}} h_{t-1} + \bm{\overline{B}} x_t, \quad
y_t = \bm{C} h_t.
\end{gathered}
\end{equation}

To improve computational efficiency, the output for the entire sequence can be computed simultaneously through a global convolution operation, enhancing both scalability and processing speed. This is formulated as:

\begin{equation}
\begin{gathered}
\bm{y} = \bm{x} \circledast \bm{\overline{K}}, \quad
\bm{\overline{K}} = (\bm{C} \bm{\overline{B}}, \bm{C} \overline{\bm{A} \bm{B}}, ..., \bm{C} \bm{\overline{A}}^{L-1} \bm{\overline{B}}),
\end{gathered}
\end{equation}

\noindent where $\circledast$ represents the convolution operation, $L$ represents the length of the sequence, and $\bm{\overline{K}}$ is the kernel derived from the SSM, specifically designed to handle sequences efficiently.
\subsubsection{Selective State Space Model}
Building upon the traditional SSM framework, the Selective SSM, called ``Mamba'', \cite{gu2023mamba} introduces a dynamic and adaptive mechanism for managing the interactions between sequential states. Unlike traditional SSM, which utilize fixed transition parameters $\mathbf{A}$ and $\mathbf{B}$, Mamba models feature input-dependent parameters, enabling a more flexible and context-aware parameterization.

In Mamba models, the parameters $\mathbf{B}$ and $\mathbf{C}$ are not static but are computed as functions of the input sequence $x$. This dynamic computation allows the model to adapt its behavior based on the specifics of the input sequence, thus providing a more nuanced understanding and processing of sequential data. The dimensions of these parameters are $\mathbf{B} \in \mathbb{R}^{B \times L \times N}$ and $\mathbf{C} \in \mathbb{R}^{B \times L \times N}$, where $B$, $L$, and $N$ denote the batch size, sequence length, and number of states, respectively. The Mamba model ensures linear scalability with respect to sequence length and demonstrates robust performance across various domains, particularly in computer vision tasks.

VMamba \cite{liu2024vmamba} and Vision Mamba (Vim) \cite{zhu2024vision} pioneered adapting Mamba for vision tasks by converting images into 2D patches and employing diverse scanning techniques before processing through Mamba blocks. VMamba uses cross-scanning, processing image patches along horizontal and vertical axes, while Vim treats images as sequences of flattened 2D patches and applies bi-directional SSM with position embeddings.

However, these initial models faced challenges in competing with full-phase ViT-based models, particularly in capturing spatial relationships and efficiently handling high-resolution images. Numerous subsequent works have been proposed to address these challenges and scanning has become an integral part of Mamba-based frameworks. It plays a crucial role in effectively capturing spatial relationships and contextual information across different parts of the input image. The following section will provide a detailed analysis of scanning techniques.

\subsection{Scanning Methods in Mamba}
\label{sec:scanning}

\begin{figure*}[!t]
  \centering
  \includegraphics[scale=0.45]{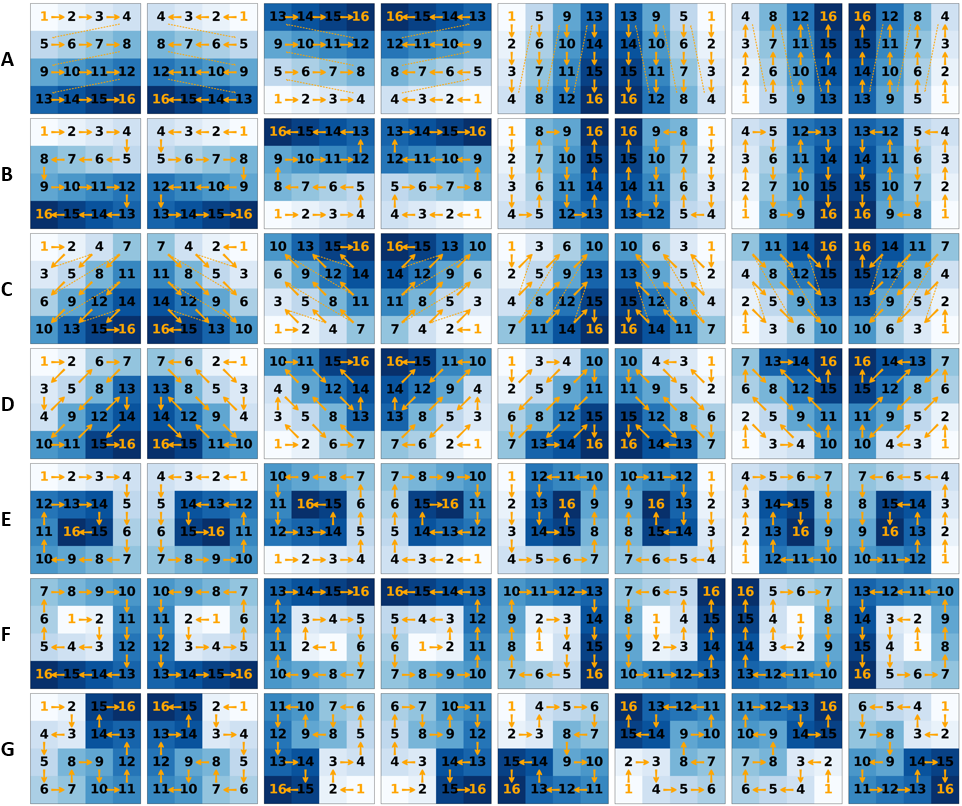}
  \caption{Various scanning patterns for image patches in Mamba are shown, with each numbered patch indicating the scanning order. A: Sequential scan without continuity, B: Sequential zigzag scan, C: Diagonal scan without continuity, D: Diagonal zigzag scan, E: Spiral scan, F: Radial scan, and G: Hilbert curve scan.}
  \label{fig:scanning}
\end{figure*}

\begin{figure*}[!t]
  \centering
  \includegraphics[scale=0.13]{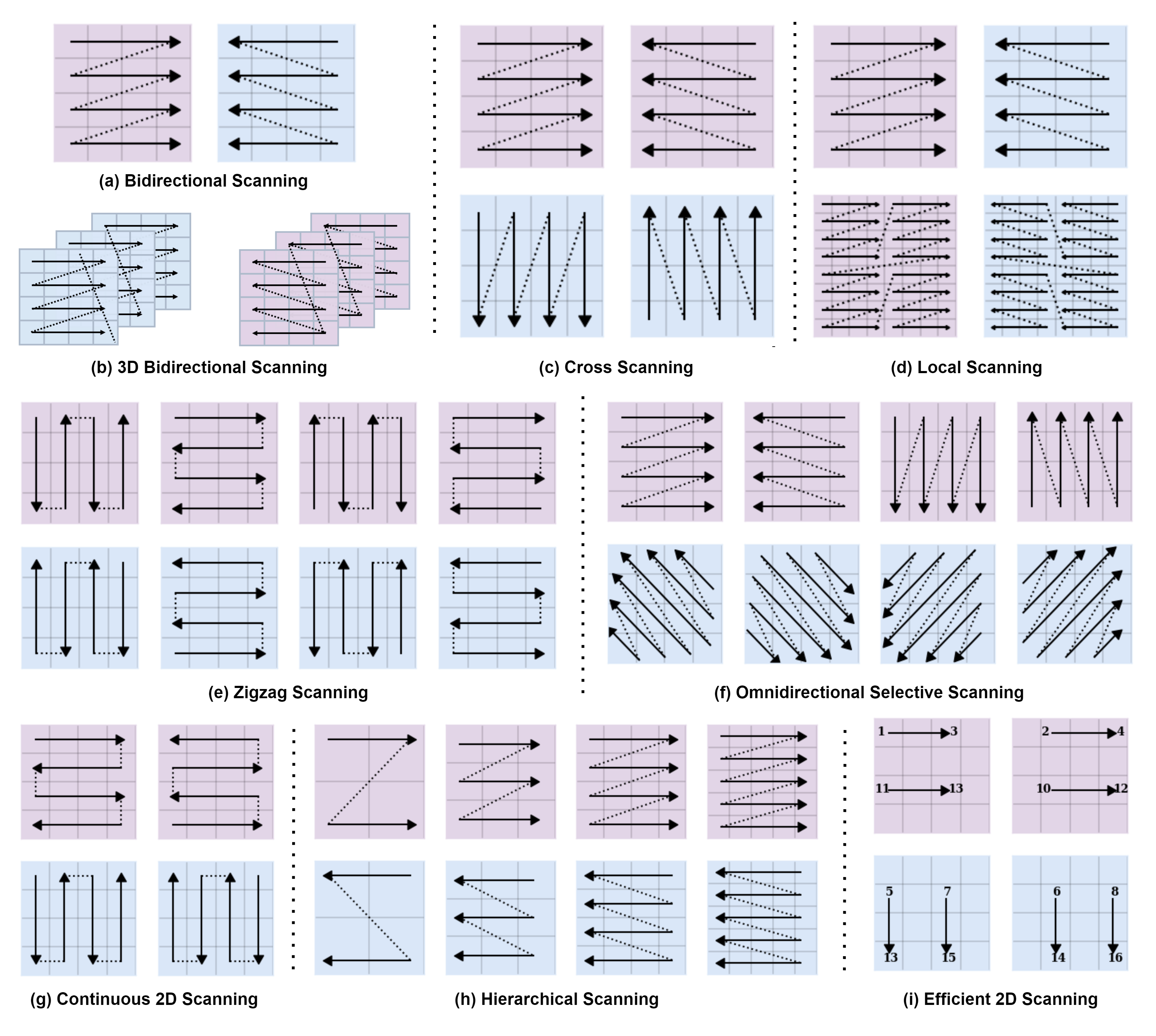}
  \caption{Different scanning methods used in Mamba models for visual tasks.}
  \label{fig:scanning_all_existing_paper}
\end{figure*}

\begin{table*}[]
\centering
\caption{Strengths and Weaknesses of Scanning Methods for Mamba Models}
\label{tab:scanning_methods_strengths_weaknesses}
\resizebox{0.9\textwidth}{!}{%

\begin{tabular}{@{}>{\centering\arraybackslash}m{3.2cm} >{\centering\arraybackslash}m{7cm} >{\centering\arraybackslash}m{7.5cm} >{\centering\arraybackslash}m{5cm}@{}}

\toprule
\textbf{Scanning Method} & \textbf{Strengths} & \textbf{Weaknesses} & \textbf{Potential Use Cases} \\ \midrule
\end{tabular}%
}
\resizebox{0.9\textwidth}{!}{%
\begin{tabular}{@{}>{\centering\arraybackslash}m{3.2cm} >{\raggedright\arraybackslash}m{8cm} >{\raggedright\arraybackslash}m{6cm} >{\raggedright\arraybackslash}m{4.5cm}@{}}
Bidirectional Scanning \cite{zhu2024vision} & 
 \textbullet\ Captures forward and backward dependencies. \newline 
\textbullet\ Enhances feature discriminability. \newline 
 \textbullet\ Reduces redundancy compared to unidirectional. & 
 \textbullet\ May introduce artifacts or resolution issues. \newline 
 \textbullet\ Increases computational complexity. & 
Object detection, image classification, and segmentation. \\ \midrule

3D Bidirectional Scanning \cite{li2024videomamba} & 
\textbullet\ Captures spatial relationships across 3D data. \newline 
\textbullet\ Processes volumetric image data effectively. & 
\textbullet\ Computationally expensive. \newline 
\textbullet\ Requires more memory for 3D volumes. & 
3D medical imaging, volumetric data, video processing. \\ \midrule

Cross Scanning \cite{qiao2024vl} & 
\textbullet\ Enhances ability to capture multi-directional spatial information. \newline 
\textbullet\ Improves performance for image data. & 
\textbullet\ Increases computational complexity. \newline 
\textbullet\ May require additional steps for the integration of cross-scanned information. & 
Text recognition, aligned object detection. \\ \midrule

Local Scanning \cite{huang2024localmamba} & 
\textbullet\ Captures fine-grained details. \newline 
\textbullet\ Flexible to combine with other methods. \newline 
\textbullet\ Efficient for processing local regions. & 
\textbullet\ May miss long-range dependencies. \newline 
\textbullet\ Requires careful tuning of window sizes. & 
Texture analysis, face recognition, and local feature extraction. \\ \midrule

Zigzag Scanning \cite{hu2024zigma} & 
\textbullet\ Provides comprehensive 2D spatial coverage. \newline 
\textbullet\ Captures both horizontal and vertical dependencies. & 
\textbullet\ May incur artifacts at scan direction changes. \newline 
\textbullet\ Computationally inefficient compared to simpler methods. & 
Remote sensing, high-resolution segmentation. \\ \midrule

Omnidirectional Selective Scanning \cite{shi2024vmambair} & 
\textbullet\ Comprehensive multi-directional scanning. \newline 
\textbullet\ Captures spatial relationships across large fields. \newline 
\textbullet\ Useful for remote sensing images. & 
\textbullet\ Increases computational complexity. \newline 
\textbullet\ Requires significant memory and processing power. & 
High-resolution remote sensing, satellite imaging, and detailed segmentation. \\ \midrule

Continuous 2D Scanning \cite{yang2024plainmamba} & 
\textbullet\ Provides smooth transitions between spatial regions. \newline 
\textbullet\ Effective for continuous data. & 
\textbullet\ Computationally intensive for large images. \newline 
\textbullet\ May not capture discrete spatial structures effectively. & 
Continuous data analysis, PlainMamba, scene segmentation. \\ \midrule

Hierarchical Scanning \cite{zhang2024motion} & 
\textbullet\ Captures both local and global features at different scales. \newline 
\textbullet\ Efficient for complex image structures. & 
\textbullet\ Increases model complexity and computational requirements. \newline 
\textbullet\ Requires careful design to balance local and global information. & 
Scene segmentation, multi-scale object detection, and medical segmentation. \\ \midrule

Efficient 2D Scanning (ES2D) / Atrous Scanning \cite{pei2024efficientvmamba} & 
\textbullet\ Reduces computational cost while maintaining the global receptive field. \newline 
\textbullet\ Captures both local and global contexts efficiently. & 
\textbullet\ May lose fine-grained details due to skip-sampling. \newline 
\textbullet\ Requires tuning of the atrous rate. & 
Mobile vision tasks, real-time detection, edge computing. \\ \midrule

2D Selective Scanning (SS2D) \cite{liu2024vmamba} & 
\textbullet\ Preserves 2D spatial dependencies. \newline 
\textbullet\ Scans in multiple directions, improving feature discriminability and enhances structural integrity &
\textbullet\ Increases computational complexity. \newline 
\textbullet\ May introduce redundancy. & 
Medical imaging, fine-grained object detection, segmentation. \\ \bottomrule

\end{tabular}%
}
\end{table*}

Scanning is a crucial process in Mamba, converting 2D visual data into 1D sequences for more efficient model processing, with various methods emerging to balance spatial integrity and computational efficiency without compromising accuracy.

Different scanning techniques serve distinct purposes in Mamba models \cite{zhu2024vision, liu2024vmamba, zhang2024motion}. For example, Local scanning \cite{huang2024localmamba} divides an image into smaller windows and processes each window independently. This method preserves local details but may fail to capture the broader context of the image. In contrast, Global scanning \cite{zhu2024vision, yang2024plainmamba} processes the entire image in one pass, which captures broader patterns but may miss finer details. Multi-head scanning \cite{hu2024zigma, shi2024vmambair} divides image patches into multiple subspaces, allowing the model to capture complex patterns while managing computational resources. On the other hand, Bidirectional scanning \cite{zhu2024vision, qiao2024vl} processes images in both horizontal and vertical directions. This approach captures spatial information effectively but requires more computational power.

To illustrate various traversal paths, Figure \ref{fig:scanning} provides a holistic overview of scanning techniques such as sequential, zigzag, spiral, radial, and hilbert curve scanning. 
Sequential scans, whether horizontal or vertical, are straightforward but may struggle with capturing long-range dependencies. Zigzag scans balance local and global information by alternating traversal direction after each row or column. Spiral and radial scans emphasize comprehensive coverage, moving from the center outward or the edges inward. They are particularly useful in applications like medical imaging and remote sensing, where detailed spatial analysis is critical. These diverse traversal paths enable Mamba models to adapt to the characteristics of different datasets and the requirements of various tasks.
These basic methods can be combined into more complex scanning approaches, as shown in Figure \ref{fig:scanning_all_existing_paper}. For example, we can derive the Omnidirectional selective scanning (Figure \ref{fig:scanning_all_existing_paper}(f)) \cite{shi2024vmambair, zhao2024rs} by combining the first, second, fifth, and seventh column from sequential scanning (Figure \ref{fig:scanning}(row A)) and third, second, eighth and fifth column from zigzag scanning (Figure \ref{fig:scanning}(row C)).

Scanning techniques can be enhanced by combining different traversal directions—local, global, or atrous—with continuous or discontinuous patterns. This adaptability optimizes the capture of both local and global features.
For instance, atrous scanning (also known as efficient or skipping scanning) uses a skipping mechanism to capture fine-grained details while maintaining computational efficiency. It provides a better spatial understanding when combined with other scanning techniques, such as sequential or zigzag.

Figure \ref{fig:scanning_all_existing_paper} represents several scanning methods that have been implemented in Mamba models, each designed with unique trade-offs. 
Vision Mamba \cite{zhu2024vision}, VL-Mamba \cite{qiao2024vl}, and Motion Mamba \cite{zhang2024motion} employ bidirectional scanning (Figure \ref{fig:scanning_all_existing_paper}(a)), traversing image patches both horizontally and vertically. This approach effectively captures a global context, though it comes with a high computational cost.
Similarly, Vivim \cite{yang2024vivim} employs spatiotemporal selective scanning, which extends 3D Bidirectional Scanning (Figure \ref{fig:scanning_all_existing_paper}(b)) \cite{li2024videomamba} by incorporating a time dimension. This approach captures both spatial and temporal features for video processing, but it is also highly computationally intensive.
VL-Mamba \cite{qiao2024vl} uses Cross-scanning (Figure \ref{fig:scanning_all_existing_paper}(c)) to capture diverse spatial features. VMamba \cite{liu2024vmamba} and VMRNN \cite{tang2024vmrnn} use the similar technique called 2D Selective Scan (SS2D) that scans from four directions toward the center.

Other methods focus on balancing local and global contexts more effectively. 
LocalMamba \cite{huang2024localmamba} and FreqMamba \cite{zhen2024freqmamba} use Local scanning (Figure \ref{fig:scanning_all_existing_paper}(d)), also known as Windowed Selective Scanning. This method divides the image into small windows and captures the local dependencies but potentially missing the global context.
ZigMa \cite{hu2024zigma} employs Zigzag scanning (Figure \ref{fig:scanning_all_existing_paper}(e)), which proceeds in a zigzag pattern to capture diverse spatial features. However, this method can be computationally demanding and complex when fusing features.
VmambaIR \cite{shi2024vmambair} and RS-Mamba \cite{zhao2024rs} utilize Omnidirectional selective scanning (Figure \ref{fig:scanning_all_existing_paper}(f)), which scans in all directions to gather comprehensive spatial information. While this method captures a broad range of features, it comes with a high computational cost.
PlainMamba \cite{yang2024plainmamba} implements Continuous 2D scanning (Figure \ref{fig:scanning_all_existing_paper}(g)), which maintains spatial continuity by sequencing adjacent tokens. However, this approach may overlook fine-grained details.
Motion Mamba \cite{zhang2024motion} utilizes Hierarchical scanning (Figure \ref{fig:scanning_all_existing_paper}(h)), capturing features at multiple levels. While this method enhances feature extraction, it also increases the model's complexity.
Finally, EfficientVMamba \cite{pei2024efficientvmamba} uses Efficient 2D scanning (ES2D), also known as Atrous selective scanning (Figure \ref{fig:scanning_all_existing_paper}(i)). This method employs skip sampling to balance global and local feature extraction, optimizing both performance and computational cost.

The effectiveness of scanning methods depends on the specific task and dataset. For instance, a study on semantic segmentation of remote sensing images \cite{mamba_scanning} found that simpler scanning strategies performed similarly to more complex ones without additional computational costs. This finding suggests that while advanced scanning techniques can capture richer spatial-temporal relationships, they do not always lead to significant performance gains. Therefore, selecting the appropriate scanning mechanism requires careful consideration of the task requirements, data characteristics, and available computational resources.
We provide distinctive characteristics with particular strengths, weaknesses, and potential use cases of existing scanning approaches as illustrated in Table \ref{tab:scanning_methods_strengths_weaknesses}. Future research could explore adaptive scanning methods that dynamically adjust based on image content, integrating learned patterns optimized during training. This direction holds the potential for enhancing the efficiency and accuracy of Mamba models across various computer
vision applications.

\section{Mamba in Computer Vision Applications}
\label{sec:mamba_in_cv}

\begin{figure}[]
  \centering
  \includegraphics[scale=0.23]{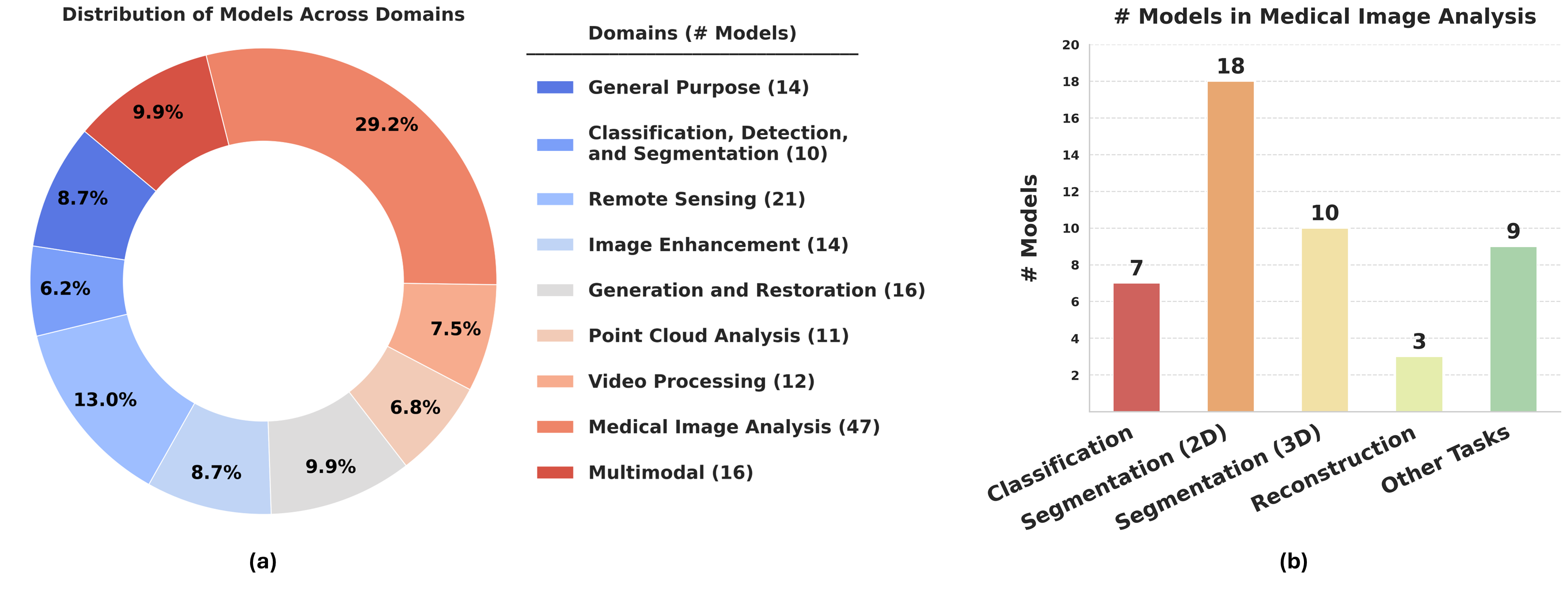}
  \caption{(a) Distribution of Mamba models across various computer vision tasks (models considered up to July 15, 2024). (b) Distribution of models specifically within medical image analysis.}
  \label{fig:paper_distribution_percentage}
\end{figure}

This section demonstrates the contributions and versatility of Mamba models in different computer vision tasks, including General Purpose Framework \ref{sec:general_purpose_V}, Image Classification, Object Detection, and Segmentation \ref{sec:classification_detection_segmentation_V}, Image Enhancement \ref{sec:image_enhancement}, Generation and Restoration \ref{sec:gen_restore_V}, 3D Point Cloud \ref{sec:point_cloud_V}, Video Processing \ref{sec:video_processing_V}, Remote Sensing \ref{sec:remote_sensing_V}, Medical Image Analysis \ref{sec:medical_image_V}, and Multimodal Models \ref{sec:multi_modals_V}. We present the distribution of Mamba models in Figure \ref{fig:paper_distribution_percentage}, which highlights their usage in various computer vision tasks.

\subsection{General Purpose}
\label{sec:general_purpose_V}

General Purpose Mamba frameworks are designed to be flexible and adaptable for classification, detection, and segmentation tasks. VMamba \cite{liu2024vmamba} enhances performance by integrating 1D scanning with 2D vision data through VSS blocks and the SS2D module, while Vision Mamba \cite{zhu2024vision} overcomes unidirectional scanning limitations using bidirectional Mamba blocks with position embeddings. Despite advancements, capturing the global context remains challenging. Vim-F \cite{zhang2024vimf} enhances the global receptive field using frequency domain information via Fast Fourier Transform (FFT), while Mamba-R \cite{wang2024mambar} reduces artifacts in feature maps with register tokens for cleaner outputs.

As models evolve, it becomes crucial to balance long-range dependency learning with computational efficiency. In order to optimize these aspects, MSVMamba \cite{shi2024multi} offers a balanced solution with a multi-scale 2D scanning method combined with a Convolutional Feed-Forward Network (ConvFFN). FractalVMamba \cite{tang2024scalable} enhances spatial relationship modeling with fractal scanning curves that adapt to varying image resolutions. LocalMamba \cite{huang2024localmamba} further optimizes these tasks by introducing windowed selective scan methods that dynamically adjust scanning strategies at different network layers, outperforming ViTs and CNNs. The challenge of balancing accuracy with computational demands is also addressed by EfficientVMamba \cite{pei2024efficientvmamba}, which integrates atrous-based selective scanning with efficient skip sampling, successfully reducing FLOPs while maintaining high performance.

\begin{table*}[]
\centering
\caption{Summary of General Purpose Mamba Models}
\label{tab:general_purpose}
\resizebox{0.9\textwidth}{!}{%
\begin{tabular}{@{}>{\centering\arraybackslash}m{3.4cm}>{\centering\arraybackslash}m{4.5cm}>{\centering\arraybackslash}m{9cm}>{\centering\arraybackslash}m{4cm}>{\centering\arraybackslash}m{1cm}@{}}

\toprule
\textbf{Models} & \textbf{Task} & \textbf{Unique Features} & \textbf{Target Domain} & \textbf{Codes} \\ \midrule

Vision Mamba \cite{zhu2024vision} \newline Jan 2024, arXiv & Classification, Detection, Segmentation & Bidirectional Mamba blocks with position embeddings & ImageNet, COCO, ADE20K & \href{https://github.com/hustvl/Vim}{\faExternalLink} \\ \midrule

VMamba \cite{liu2024vmamba} \newline Jan 2024, arXiv & Classification, Detection, Segmentation & VSS blocks with SS2D module & ImageNet, COCO, ADE20K & \href{https://github.com/MzeroMiko/VMamba}{\faExternalLink} \\ \midrule

Mamba-ND \cite{li2024mamba} \newline Feb 2024, arXiv & Classification, Action Recognition, Forecasting & Generalized Mamba architecture to arbitrary multi-dimensional data & ImageNet, HMDB-51, UCF-101, ERA5, BTCV & \href{https://github.com/jacklishufan/Mamba-ND}{\faExternalLink}  \\ \midrule

LocalMamba \cite{huang2024localmamba} \newline Mar 2024, arXiv & Classification, Detection, Segmentation & Windowed selective scan for improved local dependency capture in visual tasks & ImageNet, COCO, ADE20K & \href{https://github.com/hunto/LocalMamba}{\faExternalLink} \\ \midrule

EfficientVMamba \cite{pei2024efficientvmamba} \newline Mar 2024, arXiv & Classification, Detection, Segmentation & Selective scan with convolutional integration for lightweight vision models & ImageNet, COCO, ADE20K & \href{https://github.com/TerryPei/EfficientVMamba}{\faExternalLink} \\ \midrule

SiMBA \cite{patro2024simba} \newline Mar 2024, arXiv & Classification, Detection, Segmentation, Time-Series Forecasting & Einstein FFT for channel modeling and Mamba block for sequence modeling & ImageNet, COCO, ADE20K, Various Time-Series Datasets & NA \\ \midrule

PlainMamba \cite{yang2024plainmamba} \newline Mar 2024, arXiv & Classification, Detection, Segmentation & Non-hierarchical Mamba model with continuous 2D scanning and direction-aware updating & ImageNet, COCO, ADE20K & \href{https://github.com/ChenhongyiYang/PlainMamba}{\faExternalLink} \\ \midrule

Heracles \cite{patro2024heracles} \newline Mar 2024, arXiv & High-Resolution Image and Time-Series Analysis & Local SSM, global SSM, attention-based token interaction module & CIFAR-10, CIFAR-100, ImageNet, Oxford Flowers, Stanford Cars, COCO & \href{https://github.com/badripatro/heracles}{\faExternalLink} \\ \midrule

MambaMixer \cite{behrouz2024mambamixer} \newline Mar 2024, arXiv & Classification, Detection, Segmentation, Forecasting & Dual token and channel selection mechanism for efficient sequence modeling & ImageNet, Various Time-Series Datasets & \href{https://mambamixer.github.io/}{\faExternalLink} \\ \midrule

Mamba-R \cite{wang2024mambar} \newline May 2024, arXiv & Classification, Segmentation & Using register tokens to enhance feature map focus and token recycling & ImageNet, ADE20K & \href{https://github.com/mambar/mambar}{\faExternalLink} \\ \midrule

MSVMamba \cite{shi2024multi} \newline May 2024, arXiv & Classification, Detection, Segmentation & Multi-scale 2D scanning with ConvFFN for enhanced long-range dependency learning & ImageNet, COCO, ADE20K & \href{https://github.com/YuHengsss/MSVMamba}{\faExternalLink} \\ \midrule

FractalVMamba \cite{tang2024scalable} \newline May 2024, arXiv & Classification, Detection, Segmentation & Fractal scanning curves for improved spatial relationships & ImageNet, COCO, ADE20K & \href{https://github.com/fractalvmamba/fractalvmamba}{\faExternalLink} \\ \midrule

Vim-F \cite{zhang2024vimf} \newline May 2024, arXiv & Classification, Detection, Segmentation & FFT modeling with spatial scanning for improved global context understanding & ImageNet, COCO & \href{https://github.com/yws-wxs/Vim-F}{\faExternalLink} \\ \midrule

SUM \cite{hosseini2024sum} \newline Jun 2024, arXiv & Visual Attention Modeling & Conditional VSS (C-VSS) block integrates Mamba with U-Net & Various Image Types (5 Benchmarks) & NA \\ 

\bottomrule
\end{tabular}%
}
\end{table*}

Expanding Mamba's capabilities to high-dimensional data introduces new challenges. Mamba-ND \cite{li2024mamba} tackles these by alternating sequence orderings across dimensions, preserving the linear complexity of SSMs while achieving high accuracy in image classification and weather forecasting tasks. To enhance flexibility in image modeling, SUM \cite{hosseini2024sum} pairs the Mamba framework with a U-Net structure. Meanwhile, Heracles \cite{patro2024heracles} addresses the complexities of high-resolution images and time-series analysis by integrating local and global SSMs with attention mechanisms.

MambaMixer \cite{behrouz2024mambamixer} proposes a dual token and channel selection mechanism to improve inter- and intra-dimension communication in both vision and time series tasks. SiMBA \cite{patro2024simba} is designed to provide a simpler yet effective architecture. It integrates Einstein FFT (EinFFT) for channel modeling, setting a new standard for SSMs in both image and time-series tasks. PlainMamba \cite{yang2024plainmamba}, with its focus on spatial continuity and directional awareness, emerges as a competitive option for various visual recognition tasks. Table \ref{tab:general_purpose} illustrates an overview of General Purpose Mamba models.

\subsection{Image Classification, Object Detection, and Segmentation}
\label{sec:classification_detection_segmentation_V}

\begin{table*}[]
\centering
\caption{Summary of Mamba Models in Image Classification, Object Detection, and Segmentation}
\label{tab:image_classification_detection_segmentation}
\resizebox{0.9\textwidth}{!}{%
\begin{tabular}{@{}>{\centering\arraybackslash}m{3.2cm}>{\centering\arraybackslash}m{3.5cm}>{\centering\arraybackslash}m{10cm}>{\centering\arraybackslash}m{3.4cm}>{\centering\arraybackslash}m{1cm}@{}}

\toprule
\textbf{Models} & \textbf{Task} & \textbf{Unique Features} & \textbf{Target Domain} & \textbf{Codes} \\ \midrule

InsectMamba \cite{wang2024insectmamba} \newline Apr 2024, arXiv & Insect Pest Classification & Integration of SSMs, CNNs, MSA, and MLPs for fine-grained feature extraction & Agricultural Pest Images (5 Datasets) & NA \\ \midrule

Res-VMamba \cite{chen2024res} \newline Feb 2024, arXiv & Fine-Grained Food Classification & Residual learning within VMamba for enhanced classification accuracy & CNFOOD-241 Dataset & \href{https://github.com/ChiShengChen/ResVMamba}{\faExternalLink} \\ \midrule

Fusion-Mamba \cite{dong2024fusion} \newline Apr 2024, arXiv & Cross-Modal Object Detection & Cross-modal fusion using a hidden state space for improved feature consistency & M3FD, FLIR-Aligned Datasets & NA \\ \midrule

MiM-ISTD \cite{chen2024mim} \newline Apr 2024, arXiv & Infrared Small Target Detection & Nested structure with Outer and Inner Mamba blocks for efficient feature capture & NUAA-SIRST, IRSTD-1K Datasets & \href{https://github.com/txchen-USTC/MiM-ISTD}{\faExternalLink} \\ \midrule

SOAR \cite{verma2024soar} \newline May 2024, arXiv &  Aerial Small Object Detection & Vision Mamba-based YOLO v9 with Programmable Gradient Information (PGI) & Aerial Images (Custom Datasets) & NA \\ \midrule

Voxel Mamba \cite{zhang2024voxel} \newline Jun 2024, arXiv & 3D Object Detection & Group-free strategy for serializing 3D voxels for improved accuracy & Point Clouds (Waymo Open, nuScenes) & NA \\ \midrule

Mamba-YOLO \cite{wang2024mamba_yolo} \newline Jun 2024, arXiv & Object Detection & SSM-based YOLO with LSBlock and RGBlock for enhanced local dependencies & COCO, VOC Datasets & \href{https://github.com/HZAI-ZJNU/Mamba-YOLO}{\faExternalLink} \\ \midrule

HTD-Mamba \cite{shen2024htdmamba} \newline Jul 2024, arXiv & Hyperspectral Target Detection & Pyramid SSM for capturing multi-resolution spectral features & Hyperspectral Imaging (4 Public Datasets) & NA \\ \midrule

RWKV-SAM \cite{yuan2024mamba} \newline Jun 2024, arXiv & High-Quality Segmentation & Mixed backbone with RWKV for efficient, high-quality segmentation & High-Resolution Segmentation Datasets & NA \\ \midrule

VMamba-CS \cite{chen2024vision} \newline Jun 2024, arXiv & Crack Segmentation & VMamba-based framework for crack segmentation with high accuracy & Concrete, Asphalt, Masonry Surfaces & NA \\ 

\bottomrule
\end{tabular}%
}
\end{table*}

Mamba improves image classification, object detection, and segmentation tasks by capturing local and global features. In image classification, challenges like distinguishing pests with high camouflage and species diversity in agricultural settings have been addressed by InsectMamba \cite{wang2024insectmamba}. This model enhances accuracy by integrating SSMs with CNNs and Multi-Head Self-Attention. Similarly, Res-VMamba \cite{chen2024res} advances food image classification by combining the Mamba mechanism with deep residual learning, setting a new standard in fine-grained recognition. Additionally, Mamba models such as RSMamba \cite{chen2024rsmamba} and SpectralMamba \cite{yao2024spectralmamba} have proven effective in classifying complex remote sensing data, which will be discussed in detail in Section \ref{sec:remote_sensing_V}. Mamba’s versatility extends to medical image classification. Models like MedMamba \cite{yue2024medmamba} and MamMIL \cite{fang2024mammil} optimize feature extraction across various imaging modalities to improve classification performance and diagnostic accuracy. These advancements will be discussed in Section \ref{sec:medical_classification_V}.

In object detection, Fusion-Mamba \cite{dong2024fusion} improves cross-modal detection accuracy by mapping features into a hidden state space to reduce disparities between different modalities. However, small object detection, particularly in aerial imagery, remains difficult due to minimal data and background noise. SOAR \cite{verma2024soar} addresses this by integrating SSMs with the lightweight YOLO v9 architecture. Mamba-YOLO \cite{wang2024mamba_yolo} also builds upon the YOLO architecture by integrating SSMs with LSBlock and RGBlock modules to improve the modeling of local image dependencies for more precise detections. Moreover, MIM-ISTD \cite{chen2024mim} employs a nested Mamba architecture to improve infrared small target detection (ISTD). In 3D object detection, Voxel Mamba \cite{zhang2024voxel} uses a group-free SSM to enhance feature extraction in point cloud data. This approach overcomes the challenge of maintaining voxel spatial proximity during serialization. Additionally, HTD-Mamba \cite{shen2024htdmamba} focuses on hyperspectral data, combining a pyramid SSM with spectrally contrastive learning and spatial-encoded spectral augmentation to capture long-range dependencies and fuse multiresolution spectral features effectively.

Segmentation tasks have also seen substantial benefits from the Mamba architecture, particularly in high-resolution imagery. For example, RWKV-SAM \cite{yuan2024mamba} segments high resolution images precisely by combining Mamba with RWKV linear attention architectures. VMamba-CS \cite{chen2024vision} employed a VMamba-based encoder-decoder network that boosts the autonomous crack detection performance on construction surfaces. 
Furthermore, the Mamba architecture has revolutionized segmentation tasks in remote sensing and medical imaging, as detailed in Sections \ref{sec:remote_sensing_V} and \ref{sec:medical_segmentation_V}, respectively. Table \ref{tab:image_classification_detection_segmentation} illustrates a detailed overview of these models.

\subsection{Image Enhancement}
\label{sec:image_enhancement}

Mamba has significantly advanced image enhancement across various domains. In endoscopic imaging, exposure abnormalities often result in poor image quality. FD-Vision Mamba (FDVM-Net) \cite{zheng2024fd} addresses this by combining convolutional layers and SSMs within a C-SSM block. This approach processes phase and amplitude information separately to achieve high-quality image reconstruction. Underwater imaging faces unique issues of color distortion and blurring. PixMamba \cite{lin2024pixmamba} efficiently captures global contextual information using a dual-level architecture that enhances underwater images while managing computational costs. Similarly, WaterMamba \cite{guan2024watermamba} tackles underwater image challenges using spatial-channel omnidirectional selective scan (SCOSS) blocks to manage dependencies and model pixel and channel information flow effectively. To reduce the FLOPs, MambaUIE\&SR \cite{chen2024mambauie} integrates VSS blocks with dynamic interaction blocks.

The challenge in low-light image enhancement lies in balancing brightness with noise reduction. RetinexMamba \cite{bai2024retinexmamba} combines traditional Retinex methods with SSMs, using innovative illumination estimators and damage restorers to maintain image quality and processing speed. LLEMamba \cite{zhang2024llemamba} advances low-light enhancement by introducing a relighting-guided Mamba architecture within a deep unfolding network. This approach balances interpretability and distortion through Retinex optimization and Mamba deep priors. For single-image dehazing, UVM-Net \cite{zheng2024u} solves the challenge of handling long-range dependencies by combining local feature extraction with Bi-SSM blocks, which efficiently manage computational resources.

Mamba models also provide significant improvements in super-resolution tasks. DVMSR \cite{lei2024dvmsr} leverages Vision Mamba and Residual State Space Blocks (RSSBs) to enhance efficiency without sacrificing performance. FourierMamba \cite{li2024fouriermamba} improves image deraining by integrating Mamba into the Fourier space with zigzag coding to correlate low and high frequencies. In optical Doppler tomography, SRODT \cite{li2024sparse} enhances B-scan reconstruction accuracy by employing SSM-based learning to capture sequential and interaction information within A-scans. MLFSR \cite{gao2024mamba} and LFMamba \cite{lu2024lfmamba} apply SSM blocks to light field image super-resolution, capturing spatial and angular correlations to significantly boost performance. 

Remote sensing presents unique challenges, such as haze and low-resolution imagery. HDMba \cite{fu2024hdmba} addresses these by using window selective scan modules to capture local and global spectral-spatial information flow, thereby improving scene reconstruction in hyperspectral image dehazing. To  facilitate the evaluation of both traditional and Mamba models, BVI-RLV \cite{lin2024bvi} introduces a comprehensive dataset along with their framework for low-light video enhancement. Table \ref{tab:image_enhancement} provides a summary of the Mamba models for image enhancement.

\begin{table*}[]
\centering
\caption{Summary of Mamba Models for Image Enhancement}
\label{tab:image_enhancement}
\resizebox{0.9\textwidth}{!}{%
\begin{tabular}{@{}>{\centering\arraybackslash}m{2.9cm}>{\centering\arraybackslash}m{3.5cm}>{\centering\arraybackslash}m{10.5cm}>{\centering\arraybackslash}m{2.5cm}>{\centering\arraybackslash}m{1.8cm}@{}}

\toprule
\textbf{Models} & \textbf{Task} & \textbf{Unique Features} & \textbf{Target Domain} & \textbf{Codes} \\ \midrule

FDVM-Net \cite{zheng2024fd} \newline Feb 2024, arXiv & Exposure Correction & Frequency-domain reconstruction utilizes C-SSM blocks with separate phase and amplitude processing paths & Endoscopic Images & \href{https://github.com/zzr-idam/FDVM-Net}{\faExternalLink} \\ \midrule

UVM-Net \cite{zheng2024u} \newline Feb 2024, arXiv & Single Image Dehazing & Bi-SSM blocks combining convolutional and SSM layers for efficient long-range dependency modeling & Hazy Images & \href{https://github.com/zzr-idam/UVM-Net}{\faExternalLink} \\ \midrule

PixMamba \cite{lin2024pixmamba} \newline Jun 2024, arXiv & Image Enhancement & Dual-level architecture with EMNet and PixNet for capturing fine-grained features and ensuring global consistency & Underwater Images & \href{https://github.com/weitunglin/pixmamba}{\faExternalLink} \\ \midrule

DVMSR \cite{lei2024dvmsr} \newline May 2024, arXiv & Super-Resolution & Residual State Space Blocks (RSSBs) with distillation strategy for efficient feature extraction and reconstruction & Low-Resolution Images & \href{https://github.com/nathan66666/DVMSR}{\faExternalLink} \\ \midrule

LFMamba \cite{lu2024lfmamba} \newline Jun 2024, arXiv & Super-Resolution & SS2D mechanism for effective spatial and angular information extraction from 2D slices of 4D light fields & Light Field Images & NA \\ \midrule

WaterMamba \cite{guan2024watermamba} \newline May 2024, arXiv & Image Enhancement & SCOSS blocks for global-local information flow modeling with efficient computational complexity & Underwater Images & NA \\ \midrule

HDMba \cite{fu2024hdmba} \newline Jun 2024, arXiv & Image Dehazing & Window Selective Scan Module (WSSM) for capturing local and global dependencies in hazy regions & Hyperspectral Images & NA \\ \midrule

LLEMamba \cite{zhang2024llemamba} \newline Jun 2024, arXiv & Image Enhancement & Relighting-guided Mamba with deep unfolding network using Retinex optimization and Mamba deep priors & Low-Light Images & NA \\ \midrule

SRODT \cite{li2024sparse} \newline Apr 2024, arXiv & Optical Doppler Tomography & SSM-based sparse reconstruction with Inception-based Feedforward Network for accurate image reconstruction & Optical Doppler Images & NA \\ \midrule

MambaUIE\&SR \cite{chen2024mambauie} \newline Apr 2024, arXiv & Image Enhancement & VSS blocks and Dynamic Interaction Block for efficient global-local information synthesis & Underwater Images & \href{https://github.com/1024AILab/MambaUIE}{\faExternalLink} \\ \midrule

FourierMamba \cite{li2024fouriermamba} \newline May 2024, arXiv & Image Deraining & Fourier learning integration with spatial and channel scanning for improved frequency correlation & Rainy Images & NA \\ \midrule

RetinexMamba \cite{bai2024retinexmamba} \newline May 2024, arXiv & Image Enhancement & Retinex-based model with innovative illumination estimators and damage restoration mechanisms & Low-Light Images & NA \\ \midrule

MLFSR \cite{gao2024mamba} \newline Jun 2024, arXiv & Super-Resolution & Subspace scanning strategy and Mamba-based Global Interaction module for capturing global information & Light Field Images & NA \\ \midrule

BVI-RLV \cite{lin2024bvi} \newline Jul 2024, arXiv & Video Enhancement & Fully registered video pairs across various low-light conditions, with comprehensive benchmarks for LLVE tasks & Low-Light Videos & \href{https://doi.org/10.21227/mzny-8c77}{\faExternalLink} \\

\bottomrule
\end{tabular}%
}
\end{table*}

\subsection{Generation and Restoration}
\label{sec:gen_restore_V}

\begin{table*}[]
\centering
\caption{Summary of Mamba Models in Image Generation and Restoration}
\label{generation_restoration}
\resizebox{0.9\textwidth}{!}{%

\begin{tabular}{@{}>{\centering\arraybackslash}m{2.9cm} >{\centering\arraybackslash}m{3.5cm} >{\centering\arraybackslash}m{10.5cm} >{\centering\arraybackslash}m{3cm} >{\centering\arraybackslash}m{1cm}@{}}
\toprule
\textbf{Models} & \textbf{Task} & \textbf{Unique Features} & \textbf{Target Domain} & \textbf{Codes} \\ \midrule

DiS \cite{fei2024scalable} \newline Mar 2024, arXiv & Image Generation & SSM diffusion models for efficient long-range dependency modeling & Large Image Datasets & \href{https://github.com/feizc/DiS}{\faExternalLink} \\ \midrule

MD-Dose \cite{fu2024md} \newline Mar 2024, arXiv & Dose Prediction & Diffusion model for radiation therapy dose prediction with a Mamba noise predictor & Thoracic Tumor Data & \href{https://github.com/flj19951219/mamba\_dose}{\faExternalLink} \\ \midrule

Gamba \cite{shen2024gamba} \newline Mar 2024, arXiv & 3D Reconstruction & Single-view 3D reconstruction using Gaussian splatting and state-space integration & Objaverse, GSO Dataset & \href{https://florinshen.github.io/gamba-project/}{\faExternalLink} \\ \midrule

MotionMamba \cite{zhang2024motion} \newline Mar 2024, arXiv & Motion Generation & Hierarchical Temporal Mamba and Bidirectional Spatial Mamba blocks & HumanML3D, KIT-ML & \href{https://steve-zeyu-zhang.github.io/MotionMamba/}{\faExternalLink} \\ \midrule

ZigMa \cite{hu2024zigma} \newline Mar 2024, arXiv & Image Generation & Zigzag scanning Mamba with Stochastic Interpolant framework & FacesHQ, UCF101, CelebA-HQ, COCO & \href{https://taohu.me/zigma/}{\faExternalLink} \\ \midrule

UniTraj \cite{xu2024deciphering} \newline May 2024, arXiv & Trajectory Generation & Multi-agent trajectory generation with Spatial Masking and Bidirectional Temporal Mamba & Basketball-U, Football-U, Soccer-U & \href{https://github.com/colorfulfuture/UniTraj-pytorch}{\faExternalLink} \\ \midrule

DiM \cite{teng2024dim} \newline May 2024, arXiv & High-Resolution Image Synthesis & Multi-directional scans, learnable padding tokens, lightweight local feature enhancement & High-Resolution Images & NA \\ \midrule

DiM2 \cite{mo2024scaling} \newline May 2024, arXiv & Image and Video Generation & Scalable diffusion architecture, efficient modeling, linear complexity & Image and Video Sequences & NA \\ \midrule

DiM-3D \cite{mo2024efficient} \newline June 2024, arXiv & 3D Shape Generation & Fast inference, reduced computational demands, high-fidelity and diverse 3D shapes & 3D Point Clouds & NA \\ \midrule


Dimba \cite{fei2024dimba} \newline June 2024, arXiv & Text-to-Image Generation & Hybrid Transformer-Mamba architecture, sequentially stacked blocks, cross-attention & Text-to-Image Datasets & NA \\ \midrule

MambaIR \cite{shi2024vmambair} \newline Feb 2024, arXiv & Image Restoration & Super-resolution and denoising, Bidirectional raster scans & Natural Images & \href{https://github.com/csguoh/MambaIR}{\faExternalLink} \\ \midrule



CU-Mamba \cite{deng2024cu} \newline Apr 2024, arXiv & Image Restoration & Comprehensive image restoration using U-Net with spatial and channel SSMs & Degraded Images & NA \\ \midrule

VmambaIR \cite{shi2024vmambair} \newline Mar 2024, arXiv & Image Restoration & Omni Selective Scan (OSS) mechanism for complex image restoration tasks & Degraded Images & NA \\ \midrule

Serpent \cite{shahab2024serpent} \newline Mar 2024, arXiv & Image Restoration & Large-scale Image Restoration, Hierarchical architecture with multiple state-space models & High-Resolution Images & NA \\ \midrule

MMA \cite{cheng2024activating} \newline Mar 2024, arXiv & Image Restoration & High-definition Content Creation, Vim architecture for horizontal raster processing & Low-Resolution Images & \href{https://github.com/ArsenalCheng/MMA}{\faExternalLink} \\ \midrule 

GMSR \cite{wang2024gmsr} \newline May 2024, arXiv & Spectral Reconstruction & Gradient-guided Mamba (GM) for efficient spectral reconstruction from RGB images & RGB Images & \href{https://github.com/wxy11-27/GMSR}{\faExternalLink} \\ \midrule

\end{tabular}%
}
\end{table*}

In image generation,  the DiS model \cite{fei2024scalable} substitutes traditional U-Net-like architectures with SSM, thus minimizing computational overhead and producing high-resolution images. This capability is crucial for satellite imagery and high-definition content creation applications. Moving into medical applications, MD-Dose \cite{fu2024md} leverages Mamba’s diffusion model to simulate radiation dose distributions for cancer treatment, offering precise, patient-specific dose mappings that enhance therapy outcomes. This transition from DiS to MD-Dose showcases Mamba's versatility across fields that require detailed outputs. Similarly, Gamba \cite{shen2024gamba} combines Gaussian splatting with Mamba's state-space blocks for efficient 3D reconstruction from single-view inputs, which is particularly useful in archaeology where data availability is often limited. ZigMa \cite{hu2024zigma} further advances visual data generation with its Zigzag Mamba diffusion model and the Stochastic Interpolant framework. By improving speed and memory utilization, ZigMa generates high-quality images and videos. 

For high-resolution image synthesis, Diffusion Mamba (DiM) integrates Mamba’s efficiency with diffusion models using multi-directional scans and lightweight local feature enhancement, introducing a "weak-to-strong" training strategy for high-resolution images \cite{teng2024dim}. DiM2 refines this with scalable diffusion techniques and bidirectional SSMs, maintaining linear complexity and outperforming diffusion transformers in image and video generation \cite{mo2024scaling}. Expanding into 3D modeling, DiM-3D replaces traditional attention with Mamba architecture for diverse, high-fidelity 3D shape generation \cite{mo2024efficient}. Additionally, Dimba combines Transformer and Mamba layers for efficient text-to-image diffusion \cite{fei2024dimba}.

In image restoration, Mamba is meticulously designed to improve clarity and recover fine details. MambaIR \cite{shi2024vmambair} employs bidirectional horizontal and vertical (BD H/V) raster scans to enhance natural images, thoroughly integrating pixel data for high-resolution outputs—this method is vital for applications \cite{Rahman_Hammond_2024} where preserving fine details is important, such as in satellite imagery. CU-Mamba \cite{deng2024cu} incorporates a U-net architecture to efficiently learn and restore images with complex textures, addressing the challenges posed by intricate patterns and degraded textures. VmambaIR \cite{shi2024vmambair} introduces an OSS mechanism for processing images from multiple directions to effectively manage intricate patterns and degraded textures. The Serpent model \cite{shahab2024serpent} employs a hierarchical architecture for large-scale restoration tasks by using multiple SSMs to handle extensive datasets while reducing computational demands.

For spectral reconstruction from RGB images, GMSR \cite{wang2024gmsr} introduces Gradient-guided Mamba, which leverages Mamba's computational efficiency and global receptive field within GM blocks. This model reduces parameters and FLOPS while maintaining high accuracy to effectively address the challenges of spectral gradient attention mechanisms. Table \ref{generation_restoration} provides an overview of Mamba models in Image Generation and Restoration.

\subsection{Point Cloud Analysis}
\label{sec:point_cloud_V}

\begin{table*}[]
\centering
\caption{Summary of Mamba Models for Point Cloud Analysis}
\label{tab:3d_point_cloud}
\resizebox{0.9\textwidth}{!}{%
\begin{tabular}{@{}>{\centering\arraybackslash}m{3.5cm}>{\centering\arraybackslash}m{3.5cm}>{\centering\arraybackslash}m{9cm}>{\centering\arraybackslash}m{4cm}>{\centering\arraybackslash}m{1cm}@{}}

\toprule
\textbf{Models} & \textbf{Task} & \textbf{Unique Features} & \textbf{Target Domain} & \textbf{Codes} \\ \midrule

PointMamba \cite{liang2024pointmamba} \newline Feb 2024, arXiv & Point Cloud
Analysis & Space-filling curves for point tokenization and non-hierarchical Mamba encoder & Multiple Point Cloud Benchmarks & \href{https://github.com/LMD0311/PointMamba}{\faExternalLink} \\ \midrule

PCM \cite{zhang2024point} \newline Mar 2024, arXiv & Surface Reconstruction & CTS and positional encoding for better spatial representation & ScanObjectNN, ModelNet40, ShapeNetPart, S3DIS & \href{https://github.com/SkyworkAI/PointCloudMamba}{\faExternalLink} \\ \midrule

3DMambaIPF \cite{zhou20243dmambaipf} \newline Apr 2024, arXiv & Point Cloud Filtering & Iterative point cloud filtering with differentiable rendering loss to enhance visual realism & Small and Large-Scale Point Clouds &  NA \\ \midrule

3DMambaComplete \cite{li20243dmambacomplete} \newline Apr 2024, arXiv & Point Cloud Completion & Mamba HyperPoint Modules: Generation, Spread, and Deformation for High-Fidelity Completion & Point Cloud Completion Benchmarks &  NA \\ \midrule

Mamba3D \cite{han2024mamba3d} \newline Apr 2024, arXiv & Autonomous Driving, Robotics & Local Norm Pooling, bidirectional SSM, high efficiency and scalability  & ScanObjectNN, ModelNet40 &  \href{https://github.com/xhanxu/Mamba3D}{\faExternalLink} \\ \midrule

Point Mamba \cite{liu2024point} \newline Mar 2024, arXiv & Geo-Architectural Analysis \& Modeling & Employs octree-based strategy for structured sequencing & ModelNet40, ScanNet & \href{https://github.com/IRMVLab/Point-Mamba}{\faExternalLink} \\ \midrule

MAMBA4D \cite{liu2024mamba4d} \newline May 2024, arXiv & Point Cloud Video understanding & Disentangles space and time, Intra-frame Spatial Mamba and Inter-frame Temporal Mamba blocks & MSR-Action3D & NA \\ \midrule

OverlapMamba \cite{xiang2024overlapmamba} \newline May 2024, arXiv & LiDAR-based Place Recognition & Shift SSM with stochastic reconstruction for robust place recognition & LiDAR Range Views & NA \\ \midrule

Mamba24/8D \cite{li2024mamba24} \newline June 2024, arXiv & Point Cloud Segmentation & Multi-path serialization, ConvMamba block for local geometries & ScanNet v2, ScanNet200, nuScenes & NA \\ \midrule

PointABM \cite{chen2024pointabm} \newline June 2024, arXiv & Point Cloud Analysis & Hybrid model combining mamba and transformer for enhanced global and local feature extraction & Point Cloud Analysis Benchmarks & NA \\ \midrule

PoinTramba \cite{wang2024pointramba} \newline June 2024, arXiv & Point Cloud Analysis & Hybrid Transformer-Mamba framework with bi-directional importance-aware ordering & ScanObjectNN, ModelNet40, ShapeNetPart & \href{https://github.com/xiaoyao3302/PoinTramba}{\faExternalLink} \\ \bottomrule

\end{tabular}%
}
\end{table*}

Mamba has progressed point cloud analysis by tackling challenges such as large data volumes, unstructured data, and high computation. PointMamba \cite{liang2024pointmamba} approaches this by aligning points in a sequence, which enables efficient processing of 3D point clouds. This method reduces parameters by 44.3\% and FLOPs by 25\%, outperforming transformer-based models in accuracy. Similarly, Point Cloud Mamba (PCM) \cite{zhang2024point} enhances modeling by converting 3D point clouds into 1D sequences through Consistent Traverse Serialization (CTS) while preserving spatial adjacency. Combined with advanced positional encoding, this approach achieves SOTA performance on benchmarks such as ScanObjectNN and ModelNet40.

Addressing noise in large-scale point clouds is another crucial challenge. 3DMambaIPF \cite{zhou20243dmambaipf} introduces a differentiable rendering loss to preserve geometric detail and enhance realism in denoised structures. This model handles high-noise environments across synthetic and real-world datasets. 3DMambaComplete \cite{li20243dmambacomplete} excels in point cloud completion by transforming sparse inputs into dense outputs using a HyperPoint Generation module. It sets new benchmarks by retaining local details during reconstruction. Mamba3D \cite{han2024mamba3d} focuses on precise geometric feature extraction with its Local Norm Pooling (LNP) blocks and uses bidirectional SSMs to integrate global features. Additionally, Point Mamba \cite{liu2024point} employs an octree-based ordering system, organizing data points into a z-order curve that preserves spatial locality for effective SSM processing.

For more advanced applications, MAMBA4D \cite{liu2024mamba4d} targets 4D point cloud video understanding by disentangling spatial and temporal features with Intra-frame Spatial Mamba and Inter-frame Temporal Mamba blocks, effectively capturing long-range motion dependencies while reducing GPU memory usage.  OverlapMamba \cite{xiang2024overlapmamba} is designed for place recognition and compresses visual representations into sequences, which improves loop closure detection. PointABM \cite{chen2024pointabm} combines bidirectional SSMs with multi-head self-attention to capture comprehensive features, enhancing the analysis of point clouds. Mamba24/8D \cite{li2024mamba24} introduces a multi-path serialization strategy along with ConvMamba blocks to effectively handle long-range dependencies. Finally, PoinTramba \cite{wang2024pointramba} optimizes point cloud recognition and segmentation by integrating Transformer and Mamba architectures, using a bi-directional importance-aware ordering (BIO) strategy.
Table \ref{tab:3d_point_cloud} provides a summary of Mamba models in Point Cloud Analysis.

\subsection{Video Processing}
\label{sec:video_processing_V}

\begin{table*}[]
\centering
\caption{Summary of Mamba Models in Video Processing}
\label{tab:video_processing}
\resizebox{0.9\textwidth}{!}{%
\begin{tabular}{@{}>{\centering\arraybackslash}m{3.4cm} >{\centering\arraybackslash}m{3.5cm} >{\centering\arraybackslash}m{10cm} >{\centering\arraybackslash}m{2.5cm} >{\centering\arraybackslash}m{1cm}@{}}

\toprule
\textbf{Models} & \textbf{Task} & \textbf{Unique Features} & \textbf{Target Domain} & \textbf{Codes} \\ \midrule

ViS4mer \cite{islam2022long} \newline Nov 2022, ECCV & Long Movie Clip Classification & Combines Transformer for short spatiotemporal features and S4 for long temporal reasoning & Long Video Clips & \href{https://github.com/md-mohaiminul/ViS4mer}{\faExternalLink} \\ \midrule

VideoMamba \cite{li2024videomamba} \newline Mar 2024, arXiv & Video Understanding & Linear complexity operator, efficient long-term modeling, scalable without extensive pretraining & High-resolution Long Videos & \href{https://github.com/OpenGVLab/VideoMamba}{\faExternalLink} \\ \midrule

Video Mamba Suite \cite{chen2024video} \newline Mar 2024, arXiv & Holistic Video Understanding & Suite of 14 models, evaluated across 12 tasks, explores Mamba’s roles in video modeling & Diverse Video Understanding & \href{https://github.com/OpenGVLab/video-mamba-suite}{\faExternalLink} \\ \midrule

RhythmMamba \cite{zou2024rhythmmamba} \newline Apr 2024, arXiv & Remote Health Monitoring & Combines multi-temporal Mamba and frequency domain feed-forward for robust rPPG analysis & Facial Video Segments & \href{https://github.com/zizheng-guo/RhythmMamba}{\faExternalLink} \\ \midrule

Simba \cite{chaudhuri2024simba} \newline Apr 2024, arXiv & Skeletal Action Recognition & U-ShiftGCN with Mamba integrates spatial and temporal features & Skeleton Action Videos & NA \\ \midrule

SSM Diffusion \cite{oshima2024ssm} \newline Mar 2024, arXiv & Video Generation & Efficient SSM-based model, linear memory consumption, FVD scores for long sequences & Video Generation & \href{https://github.com/shim0114/SSM-Meets-Video-Diffusion-Models}{\faExternalLink} \\ \midrule

SSSMLV \cite{wang2023selective} \newline CVPR 2023 & Long Video Understanding & Selective S4 (S5) model, adaptive token selection, long-short masked contrastive learning & Long-form Videos & NA \\ \midrule

Matten \cite{gao2024matten} \newline May 2024, arXiv & Video Generation & Mamba-Attention architecture, spatial-temporal attention, scalability in video quality & Video Generation & NA \\ \midrule

DeMamba \cite{chen2024demamba} \newline May 2024, arXiv & AI-generated Video Detection & Plug-and-play module, enhances detectors, superior generalizability on large-scale datasets & AI-Generated Videos & \href{https://github.com/chenhaoxing/DeMamba}{\faExternalLink} \\ \midrule

VideoMambaPro \cite{lu2024videomambapro} \newline Jun 2024, arXiv & Video Understanding & Masked backpropagation with elemental residuals for SOTA action recognition & Video Action Recognition & NA \\ \midrule

SSM Event Vision \cite{zubic2024state} \newline CVPR 2024 & Event Camera Data Processing & Adaptive timescale learning, frequency variation handling, minimal high-frequency degradation & Event Camera Data & \href{https://github.com/uzh-rpg/ssms_event_cameras}{\faExternalLink}  \\ \bottomrule

\end{tabular}%
}
\end{table*}

Mamba has advanced video processing through its ability to address challenges like managing long sequences and efficiently handling high-resolution data. ViS4mer \cite{islam2022long} directly confronts the inefficiencies of self-attention mechanisms in long-range video understanding. By combining a Transformer encoder for short-range feature extraction with a multi-scale temporal S4 decoder, ViS4mer processes data over 2.5 times faster, uses 8 times less memory than pure self-attention models, and achieves SOTA accuracy in long-form video classification.

Additionally, the VideoMamba Suite \cite{chen2024video} showcases the versatility of Mamba-based models across a variety of video understanding tasks. It categorizes Mamba's applications into roles like temporal modeling, multimodal interaction, and spatial-temporal processing. These categorizations reveal Mamba’s adaptability to different video processing needs. RhythmMamba \cite{zou2024rhythmmamba} addresses the challenge of capturing quasi-periodic rPPG patterns for remote physiological measurement. In skeletal action recognition, Simba \cite{chaudhuri2024simba} integrates Mamba with U-ShiftGCN, which leads to SOTA results by enhancing both spatial and temporal modeling.

Further exploring advanced applications, Matten \cite{gao2024matten} incorporates a latent diffusion mechanism with spatial-temporal Mamba-Attention to produce high-quality videos with minimal computational cost. Meanwhile, DeMamba \cite{chen2024demamba} tackles the challenge of detecting AI-generated videos by introducing the GenVideo dataset and a detailed Mamba module, which enhances the effectiveness and reliability of detection methods across different video types.

In video understanding, VideoMambaPro \cite{lu2024videomambapro} addresses the limitations in token processing and achieves SOTA accuracy in video action recognition tasks while maintaining efficiency. SSM Event Vision \cite{zubic2024state} introduces state-space models with learnable timescales that can adapt to different inference frequencies in event-camera data. Additionally, SSM Diffusion \cite{oshima2024ssm} captures temporal dynamics for video generation, maintaining linear memory complexity while achieving competitive Frechet Video Distance (FVD) scores. Finally, Selective Structured State-Spaces (SSSMLV) \cite{wang2023selective} focuses on modeling long-term spatiotemporal dependencies in long-form videos. To improve it's effectiveness, it employs a lightweight mask generator to selectively process informative image tokens. VideoMamba \cite{li2024videomamba} also tackles video understanding by efficiently managing local redundancy and global dependencies, setting a new benchmark for video comprehension.
Table \ref{tab:video_processing} provides a summary of Mamba models used in video processing.

\subsection{Remote Sensing}
\label{sec:remote_sensing_V}

Mamba's capability of processing high-dimensional data from satellite and aerial imagery makes it suitable for Remote Sensing applications. 
To classify hyperspectral images, SpectralMamba~\cite{yao2024spectralmamba} efficiently addresses the challenges of high dimensionality and inter-band correlation by directly integrating spectral data into the classification process. HSIMamba~\cite{yang2024hsimamba} further improves upon traditional models by incorporating bidirectional processing. It helps to distinguish subtle differences in spectral signatures crucial for applications like vegetation analysis and land cover change detection. Additionally, the introduction of the 3DSS-Mamba framework \cite{he20243dss} provides a 3D-Spectral-Spatial approach, using a spectral-spatial token generation module and a novel selective scanning mechanism. RSMamba \cite{chen2024rsmamba}, SS-Mamba \cite{huang2024spectral}, and $S^2$Mamba \cite{wang2024s} enhance terrain analysis accuracy through various innovative techniques. Among them, RSMamba introduces position-sensitive dynamic multi-path activation to handle 2D non-causal data effectively and enhance classification accuracy across diverse terrains. SS-Mamba and $S^2$Mamba further refine this approach by employing spectral-spatial Mamba blocks and bi-directional scanning mechanisms, respectively, ensuring precise spatial-spectral fusion for accurate classification.
However, in semantic segmentation, Mamba-based models RS3Mamba \cite{ma2024rs3mamba} and Samba \cite{zhu2024samba} employ dual-branch networks and encoder-decoder architectures. These approaches enhance global and local data comprehension to optimize the extraction of multilevel semantic information from high-resolution images.

In pan-sharpening, merging low-resolution multi-spectral images with high-resolution panchromatic images to enhance visual details poses a significant challenge. Pan-Mamba \cite{he2024pan} addresses this issue  by utilizing channel-swapping and cross-modal Mamba blocks, which facilitate efficient data interchange between different image modalities. This method significantly enhances the spatial and spectral details to produce high-quality pan-sharpened images that preserve essential information from both input sources. Similarly, hyperspectral image denoising, which requires preserving essential details while eliminating noise, is adeptly handled by HSIDMamba \cite{liu2024hsidmamba}. It employs continuous scan blocks and bidirectional scanning to maintain the integrity of hyperspectral data.

\begin{table*}[!h]
\centering
\caption{Summary of Mamba Models in Remote Sensing}
\label{tab:remote_sensing}
\resizebox{0.9\textwidth}{!}{%
\begin{tabular}{@{}>{\centering\arraybackslash}m{3cm}>{\centering\arraybackslash}m{4cm}>{\centering\arraybackslash}m{9cm}>{\centering\arraybackslash}m{4cm}>{\centering\arraybackslash}m{1cm}@{}}

\toprule
\textbf{Models} & \textbf{Task} & \textbf{Unique Features} & \textbf{Target Domain} & \textbf{Codes} \\ \midrule

HSIMamba \cite{yang2024hsimamba} \newline Apr 2024, arXiv & Hyperspectral Image Classification & Bidirectional reversed CNN pathways for efficient spectral-spatial feature extraction & Houston 2013, Indian Pines, Pavia University & NA \\ \midrule

SpectralMamba \cite{yao2024spectralmamba} \newline Apr 2024, arXiv & Hyperspectral Image Classification & Efficient spatial-spectral encoding with dynamic masking & Satellite, Aircraft, UAV-Borne & \href{https://github.com/danfenghong/SpectralMamba}{\faExternalLink} \\ \midrule

3DSS-Mamba \cite{he20243dss} \newline May 2024, arXiv & Hyperspectral Image Classification & 3D spectral-spatial token scanning for global context modeling & Hyperspectral Images & NA \\ \midrule

RSMamba \cite{chen2024rsmamba} \newline Mar 2024, arXiv & Remote Sensing Image Classification & Dynamic multipath activation for non-causal sequence modeling & Remote Sensing Images (UC Merced, AID, RESISC45) & \href{https://github.com/KyanChen/RSMamba}{\faExternalLink} \\ \midrule

SS-Mamba \cite{huang2024spectral} \newline Apr 2024, arXiv & Hyperspectral Image Classification & Spatial-spectral token generation with feature enhancement & Hyperspectral Images & NA \\ \midrule

$S^2$Mamba \cite{wang2024s} \newline Apr 2024, arXiv & Hyperspectral Image Classification & Bidirectional spectral scanning with spatial-spectral fusion & Hyperspectral Images & \href{https://github.com/PURE-melo/S2Mamba}{\faExternalLink} \\ \midrule

HSIDMamba \cite{liu2024hsidmamba} \newline Apr 2024, arXiv & Hyperspectral Image Denoising & Bidirectional continuous scanning for enhanced spatial-spectral denoising & Hyperspectral Images & NA \\ \midrule

Pan-Mamba \cite{he2024pan} \newline Feb 2024, arXiv & Pan-sharpening & Channel swapping and cross-modal Mamba fusion for pan-sharpening & Multispectral and Panchromatic Images & \href{https://github.com/alexhe101/Pan-Mamba}{\faExternalLink} \\ \midrule

ChangeMamba \cite{chen2024changemamba} \newline Apr 2024, arXiv & Change Detection & Spatio-temporal modeling with Visual Mamba for accurate change detection & Remote Sensing Images (5 Benchmarks) & \href{https://github.com/ChenHongruixuan/MambaCD}{\faExternalLink} \\ \midrule

RSCaMa \cite{liu2024rscama} \newline Apr 2024, arXiv & Change Captioning & Joint spatial-temporal modeling with difference-aware scanning & Remote Sensing Images & \href{https://github.com/Chen-Yang-Liu/RSCaMa}{\faExternalLink} \\ \midrule

RS3Mamba \cite{ma2024rs3mamba} \newline Apr 2024, arXiv & Semantic Segmentation & Dual-branch network with collaborative global-local feature fusion & ISPRS Vaihingen, LoveDA Urban & \href{https://github.com/sstary/SSRS}{\faExternalLink} \\ \midrule

Samba \cite{zhu2024samba} \newline Apr 2024, arXiv & Semantic Segmentation & Samba blocks with UperNet for multi-level semantic extraction & LoveDA, ISPRS Vaihingen, ISPRS Potsdam & \href{https://github.com/zhuqinfeng1999/Samba}{\faExternalLink} \\ \midrule

CM-UNet \cite{liu2024cm} \newline May 2024, arXiv & Semantic Segmentation & CNN encoder with Mamba-based global-local fusion decoder & Remote Sensing Images (3 Benchmarks) & \href{https://github.com/XiaoBuL/CM-UNet}{\faExternalLink} \\ \midrule

LE-Mamba \cite{cao2024novel} \newline Apr 2024, arXiv & Image Fusion & Local-enhanced vision Mamba block with state sharing & Multispectral and Hyperspectral Images & \href{https://github.com/294coder/Efficient-MIF}{\faExternalLink} \\ \midrule

RSDehamba \cite{zhou2024rsdehamba} \newline May 2024, arXiv & Image Dehazing & Vision Dehamba Block with Direction-aware Scan for dehazing & Remote Sensing Images & NA \\ \midrule

FMSR \cite{xiao2024frequency} \newline May 2024, arXiv & Super-Resolution & Frequency-assisted Mamba for super-resolution & Remote Sensing Images & NA \\ \midrule

Seg-LSTM \cite{zhu2024seg} \newline Jun 2024, arXiv & Semantic Segmentation & xLSTM cell for semantic segmentation with encoder-decoder design & Remote Sensing Images & \href{https://github.com/zhuqinfeng1999/Seg-LSTM}{\faExternalLink} \\ \midrule

PyramidMamba \cite{wang2024pyramidmamba} \newline Jun 2024, arXiv & Semantic Segmentation & Pyramid fusion Mamba for multi-scale feature representation & Remote Sensing Images (OpenEarthMap, ISPRS Vaihingen, Potsdam) & \href{https://github.com/WangLibo1995/GeoSeg}{\faExternalLink} \\ \midrule

CDMamba \cite{zhang2024cdmamba} \newline Jun 2024, arXiv & Change Detection & Scaled residual ConvMamba for bi-temporal feature fusion & Remote Sensing Images (3 Benchmarks) & \href{https://github.com/zmoka-zht/CDMamba}{\faExternalLink} \\ \midrule

VMRNN \cite{tang2024vmrnn} \newline Mar 2024, arXiv & Spatiotemporal Forecasting & VMRNN cell integrating Mamba with LSTM for forecasting & Spatiotemporal Data & \href{https://github.com/yyyujintang/VMRNN-PyTorch}{\faExternalLink} \\ \midrule

RS-Mamba \cite{zhao2024rs} \newline Apr 2024, arXiv & Dense Prediction & Omnidirectional scanning for large spatial context modeling & Very-High-Resolution Remote Sensing Images & \href{https://github.com/walking-shadow/Official_Remote_Sensing_Mamba}{\faExternalLink} \\ 

\bottomrule
\end{tabular}%
}
\end{table*}

Furthermore, change detection, an essential aspect of remote sensing, requires precise identification of differences between multi-temporal images. ChangeMamba \cite{chen2024changemamba} and RSCaMa \cite{liu2024rscama} excel in this domain by employing different scanning mechanisms and spatial difference-guided SSM. These models effectively manage bi-temporal feature interactions and provide accurate spatial change detection. This capability is vital for monitoring environmental changes, urban development, and disaster impact assessment. Another important task named Image fusion, aiming to integrate spatial and spectral data without losing fine details, is addressed by LE-Mamba \cite{cao2024novel}. It uses Mamba blocks within U-shaped networks and locally enhanced vision Mamba blocks, respectively. To tackle the limitations of large-scale images and object variations, CM-UNet \cite{liu2024cm} combines CNN-based encoders with Mamba-based decoders to enhance global-local information fusion. Furthermore, RSDehamba \cite{zhou2024rsdehamba} integrates the SSM framework into a U-Net architecture for remote sensing image dehazing, while FMSR \cite{xiao2024frequency} employs a multi-level fusion architecture for super-resolution. Models like Seg-LSTM \cite{zhu2024seg}, PyramidMamba \cite{wang2024pyramidmamba}, and CDMamba \cite{zhang2024cdmamba} refine multi-scale feature representation to enhance segmentation accuracy and change detection precision. Lastly, VMRNN \cite{tang2024vmrnn} integrates Vision Mamba blocks with LSTM to balance efficiency and accuracy in spatiotemporal prediction tasks. Table \ref{tab:remote_sensing} illustrates a summary of Mamba models in Remote Sensing.

\subsection{Medical Image Analysis}
\label{sec:medical_image_V}
Mamba is tremendously popular for medical image analysis. This section explores Mamba models for different medical image analysis tasks such as classification, segmentation, and reconstruction.

\subsubsection{Medical Image Classification}
\label{sec:medical_classification_V}


\begin{table*}[]
\centering
\caption{Summary of Mamba Models for Medical Image Classification}
\label{tab:med_classification}
\resizebox{0.9\textwidth}{!}{%
\begin{tabular}{@{}>{\centering\arraybackslash}m{3.5cm}>{\centering\arraybackslash}m{3cm}>{\centering\arraybackslash}m{9cm}>{\centering\arraybackslash}m{3cm}>{\centering\arraybackslash}m{2cm}@{}}

\toprule
\textbf{Models} & \textbf{Task} & \textbf{Unique Features} & \textbf{Target Domain} & \textbf{Codes} \\ \midrule

MedMamba \cite{yue2024medmamba} \newline Mar 2024, arXiv & Multimodal Image Classification & SS-Conv-SSM block, grouped convolution, efficient processing & CT, MRI, X-ray, Ultrasound & \href{https://github.com/YubiaoYue/MedMamba}{\faExternalLink} \\ \midrule

BU-Mamba \cite{nasiri2024vision} \newline Jul 2024, arXiv & Breast Ultrasound Classification & Comparison with CNNs and ViTs, significant results & Breast Ultrasound Imaging & \href{https://github.com/anasiri/BU-Mamba}{\faExternalLink} \\ \midrule

MamMIL \cite{fang2024mammil} \newline Mar 2024, arXiv & WSI Classification & Bidirectional SSM, 2D context-aware block & Pathology, Cancer Diagnosis & NA \\ \midrule

MambaMIL \cite{yang2024mambamil} \newline Mar 2024, arXiv & WSI Classification & Sequence Reordering, long sequence modeling & Computational Pathology & \href{https://github.com/isyangshu/MambaMIL}{\faExternalLink} \\ \midrule

Vim4Path \cite{nasiri2024vim4path} \newline Mar 2024, arXiv & Histopathology Classification & Vision Mamba in DINO framework, patch/slide-level & Histopathology & \href{https://github.com/AtlasAnalyticsLab/Vim4Path}{\faExternalLink} \\ \midrule

CMViM \cite{yang2024cmvim} \newline Mar 2024, arXiv & 3D Multimodal Classification & Masked Vim Autoencoder, contrastive learning & Alzheimer's Disease, MRI/PET & NA \\ \midrule

Vision Mamba \cite{gurung2024vision} \newline Jun 2024, arXiv & Alzheimer's Classification & Dynamic states, selective scan for 3D MRI & Alzheimer's Disease, 3D MRI & NA \\ \bottomrule

\end{tabular}%
}
\end{table*}

MedMamba \cite{yue2024medmamba} addresses the challenges of CNNs and ViTs by integrating convolutional layers with SSMs. Building upon the strengths of SSMs, BU-Mamba \cite{nasiri2024vision} applies Vision Mamba to breast ultrasound image classification. It achieves better performance, especially when working with limited data. However, classifying Whole Slide Images (WSIs) is challenging due to their gigapixel scale, which complicates efficient feature extraction. To address this, MamMIL \cite{fang2024mammil} integrates bidirectional SSMs with a 2D context-aware block to improve WSIs feature detection. It preserves spatial relationships while also reducing memory usage. MambaMIL \cite{yang2024mambamil} further optimizes tissue sample arrangement with the Sequence Reordering Mamba (SR-Mamba) by improving feature extraction while reducing overfitting. Vim4Path \cite{nasiri2024vim4path} takes WSI analysis a step further, using self-supervised learning with the DINO framework to significantly improve feature encoding.

For 3D medical imaging, CMViM \cite{yang2024cmvim} improves multimodal data integration by using a contrastive masked Vim autoencoder. The model refines representations across imaging modalities and increases diagnostic accuracy for Alzheimer’s disease. Vision Mamba \cite{gurung2024vision} enhances 3D MRI processing by employing dynamic state representations and a selective scan algorithm, which efficiently captures spatial information and improves early detection accuracy. Table \ref{tab:med_classification} provides a detailed summary of Mamba models used in medical image classification.

\subsubsection{Medical Image Segmentation} 
\label{sec:medical_segmentation_V}


\begin{table*}[]
\centering
\caption{Summary of Mamba Models for 2D Medical Image Segmentation}
\label{tab:med_2d_segmentation}
\resizebox{0.9\textwidth}{!}{%
\begin{tabular}{@{}>{\centering\arraybackslash}m{3.75cm}>{\centering\arraybackslash}m{3.6cm}>{\centering\arraybackslash}m{9.5cm}>{\centering\arraybackslash}m{3cm}>{\centering\arraybackslash}m{1cm}@{}}
\toprule
\textbf{Model} & \textbf{Task (Type)} & \textbf{Unique Features} & \textbf{Target Domain} & \textbf{Code} \\ \midrule

U-Mamba \cite{ma2024u} \newline Jan 2024, arXiv & CT, MRI, Endo., Micro. Segmentation (2D) & Hybrid CNN-SSM, encoder-decoder structure, self-configuring mechanism & 3D CT, MR, Endo., Micro. & \href{https://wanglab.ai/u-mamba.html}{\faExternalLink} \\ \midrule

Mamba-UNet \cite{wang2024mamba} \newline Feb 2024, arXiv & CT \& MRI Segmentation (2D) & Symmetrical encoder-decoder, Visual Mamba bottleneck, hierarchical VSS blocks & MRI, CT Abdomen & \href{https://github.com/ziyangwang007/Mamba-UNet}{\faExternalLink} \\ \midrule

VM-UNet \cite{ruan2024vm} \newline Feb 2024, arXiv & Abdominal \& Skin Lesion Segmentation (2D) & Asymmetrical encoder-decoder, hierarchical VSS blocks, feature fusion & ISIC17, ISIC18, Synapse & \href{https://github.com/jcruan519/vm-unet}{\faExternalLink} \\ \midrule

Swin-UMamba \cite{liu2024swin} \newline Feb 2024, arXiv & Endo., \& Micro., MRI Segmentation (2D) & Hybrid Transformer-Mamba, Swin transformer pretraining, Mamba-based decoding & Abdomen MRI, Endo., Micro. & \href{https://github.com/JiarunLiu/Swin-UMamba}{\faExternalLink} \\ \midrule

VM-UNet V2 \cite{zhang2024vm} \newline Mar 2024, arXiv & Polyp And Skin Lesion Segmentation (2D) & Enhanced VSS encoder-decoder, Semantics and Detail Infusion (SDI) module & ISIC17, ISIC18, CVC-ClinicDB & \href{https://github.com/nobodyplayer1/VM-UNetV2}{\faExternalLink} \\ \midrule

H-vmunet \cite{wu2024h} \newline Mar 2024, arXiv & Skin, Spleen, Endo. Segmentation (2D) & High-Order Vision Mamba, selective scanning for local feature learning & ISIC17, Spleen, CVC-ClinicDB & \href{https://github.com/wurenkai/H-vmunet}{\faExternalLink} \\ \midrule

UltraLight VM-UNet \cite{wu2024ultralight} \newline Mar 2024, arXiv & Lightweight Skin Lesion Segmentation (2D) & Lightweight design for high segmentation accuracy in resource-limited environments & Skin Lesion Datasets & \href{https://github.com/wurenkai/UltraLight-VM-UNet}{\faExternalLink} \\ \midrule

P-Mamba \cite{ye2024p} \newline Feb 2024, arXiv & Pediatric Echo. Segmentation (2D) & Dual-Branch DWT-PMD, Perona-Malik Diffusion blocks, Mamba layers for noise suppression & Pediatric Echocardiography & N/A \\ \midrule

Weak-Mamba-UNet \cite{wang2024weak} \newline Feb 2024, arXiv & Weakly-Supervised MRI Segmentation (2D) & Weakly supervised CNN, ViT, Mamba, cross-supervision with sparse annotations & MRI Cardiac Segmentation & \href{https://github.com/ziyangwang007/Weak-Mamba-UNet}{\faExternalLink} \\ \midrule

Semi-Mamba-UNet \cite{ma2024semi} \newline Feb 2024, arXiv & Semi-Supervised MRI Segmentation (2D) & Semi-supervised, self-supervised contrastive learning, pseudo-labels for segmentation & MRI, CT & \href{https://github.com/ziyangwang007/mamba-unet}{\faExternalLink} \\ \midrule

ProMamba \cite{xie2024promamba} \newline Mar 2024, arXiv & Accurate Polyp Segmentation (2D) & Vision Mamba with prompting, box prompt integration, Vision Mamba feature extraction & Colonoscopy Datasets & N/A \\ \midrule

MUCM-Net \cite{yuan2024mucm} \newline May 2024, arXiv & Mobile Skin Lesion Segmentation (2D) & Mamba-UCM Layer, mobile deployment, high accuracy with low computational needs & ISIC Datasets & \href{https://github.com/chunyuyuan/MUCM-Net}{\faExternalLink} \\ \midrule

AC-MambaSeg \cite{nguyen2024ac} \newline May 2024, arXiv & Enhanced Skin Lesion Segmentation (2D) & Hybrid CNN-Mamba, Convolutional Block Attention Module, Selective Kernel Bottleneck & ISIC-2018, PH2 & \href{https://github.com/vietthanh2710/AC-MambaSeg}{\faExternalLink} \\ \midrule

ViM-UNet \cite{archit2024vim} \newline Apr 2024, arXiv & Microscopy Instance Segmentation (2D) & Vision Mamba architecture, global field of view, higher efficiency compared to UNETR & Microscopy Segmentation & \href{https://github.com/constantinpape/torch-em/blob/main/vimunet.md}{\faExternalLink} \\ \midrule

HC-Mamba \cite{xu2024hc} \newline May 2024, arXiv & Organ And Skin Lesion Segmentation (2D) & Dilated and depthwise separable convolutions, low computational cost for efficient processing & Synapse, ISIC17, ISIC18 & N/A \\ \midrule

SliceMamba \cite{fan2024slicemamba} \newline Jul 2024, arXiv & Skin Lesion And Polyp Segmentation (2D) & Bidirectional Slice Scan module, locally sensitive feature segmentation for detailed analysis & Skin Lesion, Polyp Datasets & N/A \\ \midrule

MHS-VM \cite{ji2024mhs} \newline Jun 2024, arXiv & Medical Image Segmentation (2D) & Multi-Head Scan, Scan Route Attention, better performance with fewer parameters & Skin Lesion, Multi-Organ & \href{https://github.com/PixDeep/MHS-VM}{\faExternalLink} \\ \midrule

xLSTM-UNet \cite{chen2024xlstm} \newline Jul 2024, arXiv & Comprehensive Medical Image Analysis (2D\&3D) & Vision-LSTM (xLSTM) backbone, local feature extraction, long-range dependency capturing & MRI, Endoscopy, Microscopy & \href{http://tianrun-chen.github.io/xLSTM-UNet/}{\faExternalLink} \\ \midrule 

\end{tabular}%
}
\end{table*}


\begin{table*}[]
\centering
\caption{Summary of Mamba Models for 3D Medical Image Segmentation}
\label{tab:med_3d_segmentation}
\resizebox{0.9\textwidth}{!}{%
\begin{tabular}{@{}>{\centering\arraybackslash}m{3.5cm}>{\centering\arraybackslash}m{4cm}>{\centering\arraybackslash}m{9cm}>{\centering\arraybackslash}m{2.8cm}>{\centering\arraybackslash}m{1cm}@{}}
\toprule
\textbf{Models} & \textbf{Task (Type)} & \textbf{Unique Features} & \textbf{Target Domain} & \textbf{Codes} \\ \midrule

LightM-UNet \cite{liao2024lightm} \newline Mar 2024, arXiv & Lightweight Medical Segmentation (3D) & Residual Vision Mamba Layers, hierarchical spatial modeling, efficient lightweight structure & 2D/3D Medical Datasets & \href{https://github.com/MrBlankness/LightM-UNet}{\faExternalLink} \\ \midrule

LKM-UNet \cite{wang2024large} \newline Mar 2024, arXiv & Spatial Modeling For Medical Images (3D) & Large Mamba kernels, location-aware sequence modeling, hierarchical Mamba blocks & Synapse, ISIC18 & \href{https://github.com/wjh892521292/LKM-UNet}{\faExternalLink} \\ \midrule

TM-UNet \cite{tang2024rotate} \newline Mar 2024, arXiv & Robust Medical Segmentation (3D) & Triplet SSM, Triplet-SSM bottleneck layer, residual connections for robust segmentation & ISIC17, ISIC18, CVC-ClinicDB & N/A \\ \midrule

Mamba-HUNet \cite{shahriar2024integrating} \newline Mar 2024, arXiv & Feature Extraction For Medical Images (3D) & Hierarchical Upsampling, VSS blocks, patch merging layers for efficient feature extraction & MRI Scans & N/A \\ \midrule

SegMamba \cite{xing2024segmamba} \newline Jan 2024, arXiv & Detailed 3D Volume Feature Modeling (3D) & Tri-Orientated Mamba, gated spatial convolution for detailed volume feature modeling & BraTS2023 & \href{https://github.com/ge-xing/SegMamba}{\faExternalLink} \\ \midrule

nnMamba \cite{gong2024nnmamba} \newline Feb 2024, arXiv & Dense Prediction, 3D Landmark Detection & Mamba-in-Convolution with Channel-Spatial Siamese learning, channel-scaling and sequential learning for dense prediction & 3D Medical Datasets & \href{https://github.com/lhaof/nnMamba}{\faExternalLink} \\ \midrule

T-Mamba \cite{hao2024t} \newline Apr 2024, arXiv & Tooth CBCT Segmentation (3D) & Frequency-Spatial Integration, gated unit, frequency-domain features for better extraction & 2D X-Ray, 3D CBCT & \href{https://github.com/isbrycee/T-Mamba}{\faExternalLink} \\ \midrule

Vivim \cite{yang2024vivim} \newline Jan 2024, arXiv & Medical Video Segmentation (3D) & Temporal Mamba Blocks, temporal compression, boundary-aware scanning for dynamic cues & Thyroid, Breast, Polyp Videos & \href{https://github.com/scott-yjyang/Vivim}{\faExternalLink} \\ \midrule

TokenUnify \cite{chen2024tokenunify} \newline May 2024, arXiv & EM Neuron Segmentation (3D) & Random, next, and next-all token prediction for scalable autoregressive visual pre-training & EM Neuron Segmentation & \href{https://github.com/ydchen0806/TokenUnify}{\faExternalLink} \\ \midrule

CAF-MambaSegNet \cite{khan2024convolution} \newline Jun 2024, arXiv & Cardiac Segmentation (3D) & Mamba-based Channel-Spatial Aggregator and Linear Factorized Mamba Block. & Cardiac Imaging & N/A \\ \bottomrule

\end{tabular}%
}
\end{table*}

U-Mamba \cite{ma2024u} is one of the early approaches to use SSMs in medical image segmentation. It employs a hybrid CNN-SSM block structure within a U-Net framework to capture long-range dependencies, which traditional methods often struggle to handle. Leveraging the strengths of SSMs, Mamba-UNet \cite{wang2024mamba} and VM-UNet \cite{ruan2024vm} offer refinements through symmetrical and asymmetrical encoder-decoder structures. These models perform better across various medical image datasets, including abdominal and skin lesion imaging.

To further improve performance, Swin-UMamba \cite{liu2024swin} combined Mamba with the hierarchical attention mechanisms of Swin Transformers to obtain the benefits from ImageNet pre-training. VM-UNet V2 \cite{zhang2024vm} introduced a Semantics and Detail Infusion (SDI) mechanism to optimize feature fusion, while H-vmunet \cite{wu2024h} addressed redundancy in long-range feature extraction through selective scanning. LightM-UNet \cite{liao2024lightm} prioritized computational efficiency by employing residual Vision Mamba layers. UltraLight VM-UNet \cite{wu2024ultralight} significantly reduced parameters without compromising accuracy, and LMa-UNet \cite{wang2024large} utilized large-window SSM blocks for efficient long-range feature capture. To improve spatial-channel integration, TM-UNet \cite{tang2024rotate} incorporated Triplet-SSM modules, and Mamba-HUNet \cite{shahriar2024integrating} focused on hierarchical upsampling. P-Mamba \cite{ye2024p} addressed noise reduction and efficient feature extraction using a dual-branch framework with DWT-based Perona-Malik Diffusion blocks.

In weakly supervised learning, Weak-Mamba-UNet \cite{wang2024weak} explores the combination of CNNs, ViTs, and Mamba architectures to address the challenges of scribble-based annotations. Semi-Mamba-UNet \cite{ma2024semi} enhances feature learning from unlabeled data through self-supervised contrastive learning. ProMamba \cite{xie2024promamba} specializes in polyp segmentation by integrating Vision Mamba with prompt technologies. MUCM-Net \cite{yuan2024mucm} and AC-MambaSeg \cite{nguyen2024ac} target skin lesion segmentation, combining Mamba with advanced feature extraction techniques. While MUCM-Net optimizes the Mamba layer to emphasize mobile deployment, AC-MambaSeg focuses on improving feature extraction and background noise suppression using a CBAM-based attention mechanism \cite{rahman2020efficient}. For microscopic instance segmentation, ViM-UNet \cite{archit2024vim} provides a global field of view with greater efficiency. Addressing the challenges of reduced resolution and information loss, HC-Mamba \cite{xu2024hc} employs dilated and depthwise separable convolutions.To improve local feature modeling, SliceMamba \cite{fan2024slicemamba} introduces a Bidirectional Slice Scan module, while xLSTM-UNet \cite{chen2024xlstm} incorporates Vision-LSTM to capture long-range dependencies and outperforms traditional segmentation models.

3D medical image segmentation presents unique challenges due to the complexity and volume of data. SegMamba \cite{xing2024segmamba} addresses this with its Tri-Orientated Mamba (ToM) module, which extracts features from axial, sagittal, and coronal planes for comprehensive anatomical coverage. LightM-UNet \cite{liao2024lightm} improves efficiency by integrating residual Vision Mamba layers, making it well-suited for clinical environments. LKM-UNet \cite{wang2024large} uses large-kernel SSM blocks to model long-range spatial dependencies, outperforming small-window transformers and conventional CNNs. nnMamba \cite{gong2024nnmamba} strengthens segmentation by combining CNNs and SSMs through the MICCSS block. This combination enhances spatial and channel relationship modeling and help to  improve landmark detection, classification, and 3D segmentation tasks. T-Mamba \cite{hao2024t} steps into more complex modalities by incorporating frequency-domain features, which enhance segmentation quality. In medical video object segmentation, Vivim \cite{yang2024vivim} ensures boundary coherence across frames through temporal Mamba blocks.
Additionally, TokenUnify \cite{chen2024tokenunify}, and CAF-MambaSegNet \cite{khan2024convolution} introduce innovative 2D segmentation solutions. It addresses cumulative errors in visual autoregression through random token prediction, leveraging the Mamba for efficient long-sequence modeling. CAF-MambaSegNet eliminates traditional convolution and self-attention mechanisms, using Channel Aggregator and Spatial Aggregator modules to independently extract features and achieve effective segmentation with reduced computational complexity. Tables \ref{tab:med_2d_segmentation} and \ref{tab:med_3d_segmentation} present the overview of Mamba models for 2D and 3D medical image segmentation, respectively.

\subsubsection{Medical Image Reconstruction}
\label{sec:medical_image_reconstration_V}
In medical image reconstruction, converting raw data from MRI, CT, and PET into high-quality images is crucial for accurate diagnosis and treatment. However, the process faces challenges, including noise, artifacts, and the need for computational efficiency.
To address these issues, MambaMIR \cite{huang2024mambamir} was developed for fast MRI and SVCT tasks. It enhances Monte Carlo-based uncertainty estimation and reduces noise and artifacts by using an arbitrary mask mechanism. While this approach improved image clarity, the need for further refinement led to the creation of MambaMIR-GAN \cite{huang2024mambamir}. This variant incorporates adversarial training to sharpen images and enhance visual quality.

The challenge of integrating Multimodal MRI data for better reconstruction led to the development of MMR-Mamba \cite{zou2024mmr}. This model combines spatial and frequency domain information using the Target modality-guided Cross Mamba (TCM) module for spatial data integration and the Selective Frequency Fusion (SFF) module for recovering high-frequency details. The Adaptive Spatial-Frequency Fusion (ASFF) module also integrates data across both domains, providing a robust solution for Multimodal MRI reconstruction. Table \ref{tab:med_reconstruction} shows the overview of Mamba models in medical image reconstruction.

\begin{table*}[]
\centering
\caption{Summary of Mamba Models for Medical Image Reconstruction}
\label{tab:med_reconstruction}
\resizebox{0.9\textwidth}{!}{%
\begin{tabular}{@{}>{\centering\arraybackslash}m{4cm}>{\centering\arraybackslash}m{4cm}>{\centering\arraybackslash}m{9cm}>{\centering\arraybackslash}m{2.8cm}>{\centering\arraybackslash}m{1cm}@{}}

\toprule
\textbf{Models} & \textbf{Task} & \textbf{Unique Features} & \textbf{Target Domain} & \textbf{Codes} \\ \midrule

MambaMIR \cite{huang2024mambamir} \newline arXiv - Feb 2024 & Reconstruction and Uncertainty Estimation & Arbitrary-mask mechanism, global receptive fields, uncertainty estimation & MRI, SVCT (Knee, Chest, Abdomen) & \href{https://github.com/jiahaohuang/MambaMIR}{\faExternalLink} \\ \midrule

MambaMIR-GAN \cite{huang2024mambamir} \newline arXiv - Feb 2024 & Reconstruction and Uncertainty Estimation & GAN-based variant, dynamic weights, enhanced uncertainty mapping & MRI, SVCT (Knee, Chest, Abdomen) & \href{https://github.com/jiahaohuang/MambaMIR-GAN}{\faExternalLink} \\ \midrule

MMR-Mamba \cite{zou2024mmr} \newline arXiv - Jun 2024 & MRI Reconstruction & TCM, SFF, and ASFF modules fusion & Multimodal MRI & \href{https://github.com/jingzou/MMR-Mamba}{\faExternalLink} \\ \bottomrule

\end{tabular}%
}
\end{table*}

\subsubsection{Other Tasks in Medical Imaging}
\label{sec:medical_other_tasks_V}

Mamba has driven significant advancements in various medical imaging tasks beyond traditional applications such as classification and segmentation.
For example, VMambaMorph \cite{wang2024vmambamorph} tackles complex Multimodality image alignment using a hybrid VMamba-CNN network, but handling intricate structures remains challenging. Similarly, the Motion-Guided Dual-Camera Tracker (MGDC Tracker) \cite{zhang2024motion2} improves endoscopy tracking through a cross-camera template strategy and Mamba-based motion prediction.  However, maintaining consistent accuracy across varying environments remains a challenge.

In radiotherapy, MD-Dose \cite{fu2024md} enhances radiotherapy dose prediction with a Mamba-based diffusion model. this approach improves precision but requires better anatomical integration. Meanwhile, BI-Mamba \cite{yang2024cardiovascular} reduces radiation exposure in cardiovascular risk prediction by capturing long-range dependencies in chest X-rays. Despite these benefits, achieving high accuracy with lower-resolution imaging still remains a challenge.

Both I2I-Mamba \cite{atli2024i2i} and VM-DDPM \cite{ju2024vm} advance Multimodal image synthesis by integrating Mamba blocks within convolutional and diffusion models. However, they still face challenges when scaling across datasets of varying sizes. Moreover, SMamba-UNet \cite{ji2024self} and Deform-Mamba \cite{ji2024deform} improve image resolution by leveraging self-prior-guided networks, but capturing fine-grained details remains a complex task.

Finally, SR-Mamba \cite{cao2024srmamba} advances surgical phase recognition by capturing long-distance temporal relationships in videos. It simplifies training and enhances accuracy but requires further refinement for consistent performance across different surgical procedures.

\subsection{Multimodal}
\label{sec:multi_modals_V}

\begin{table*}[]
\centering
\caption{Summary of Multimodal Mamba Models}
\label{tab:multi_modals}
\resizebox{0.9\textwidth}{!}{%
\begin{tabular}{@{}>{\centering\arraybackslash}m{3cm} >{\centering\arraybackslash}m{3.5cm} >{\centering\arraybackslash}m{9.5cm} >{\centering\arraybackslash}m{3.5cm} >{\centering\arraybackslash}m{1cm}@{}}

\toprule
\textbf{Models} & \textbf{Task} & \textbf{Unique Features} & \textbf{Target Domain} & \textbf{Codes} \\ \midrule

SurvMamba \cite{chen2024survmamba} \newline Apr 2024, arXiv & Survival Prediction & Hierarchical Mamba and Interaction Fusion Mamba for Multimodal fusion & Pathological and Genomic Data & NA \\ \midrule

Meteor \cite{lee2024meteor} \newline May 2024, arXiv & Visual Language Understandings & Rationale traversal Mamba and multi-faceted rationale embedding for LLVMs & Vision and Language Data & \href{https://github.com/ByungKwanLee/Meteor}{\faExternalLink} \\ \midrule 

TransMA \cite{wu2024transma} \newline Jul 2024, arXiv & mRNA Delivery
Prediction & Multimodal molecular structure fusion and mol-attention mechanism & Molecular Structure Data & \href{https://github.com/wklix/TransMA}{\faExternalLink} \\ \midrule

CMViM \cite{yang2024cmvim} \newline Mar 2024, arXiv & Alzheimer's Diagnosis & Contrastive Masked Vim Autoencoder with Intra- and Inter-Modal Learning & 3D Medical Images & NA \\ \midrule

SpikeMba \cite{li2024spikemba} \newline Apr 2024, arXiv & Video Content Grounding & Multimodal spiking saliency detector and SSM-based contextual reasoning & Video and Language Data & NA \\ \midrule

Broad Mamba \cite{shou2024revisiting} \newline Apr 2024, arXiv & Emotion Recognition & Broad Mamba for sequence modeling and probability-guidance fusion strategy & Conversational Emotion Data & NA \\ \midrule

Mamba-FETrack \cite{huang2024mamba} \newline Apr 2024, arXiv & RGB-Event Tracking & Modality-specific Mamba backbones and efficient interactive learning & RGB and Event Data & \href{https://github.com/Event-AHU/Mamba_FETrack}{\faExternalLink} \\ \midrule

MambaTalk \cite{xu2024mambatalk} \newline Mar 2024, arXiv & Gesture Synthesis & Two-stage modeling with discrete motion priors and multimodal integration & Gesture Data & NA \\ \midrule

VL-Mamba \cite{qiao2024vl} \newline Mar 2024, arXiv & Multimodal Learning & SSMs for long-sequence modeling and vision selective scan mechanism & Multimodal Language Models & \href{https://github.com/ZhengYu518/VL-Mamba}{\faExternalLink} \\ \midrule

MambaMorph \cite{wang2024vmambamorph} \newline Apr 2024, arXiv & Image Registration & Visual SSM with cross-scan module and hybrid VMamba-CNN network & 3D Medical Images & \href{https://github.com/ziyangwang007/VMambaMorph}{\faExternalLink} \\ \midrule

Sigma \cite{wan2024sigma} \newline Apr 2024, arXiv & Multimodal Segmentation & Siamese Mamba network and Mamba fusion mechanism for modality interaction & RGB-Thermal, RGB-Depth Data & \href{https://github.com/zifuwan/Sigma}{\faExternalLink} \\ \midrule

FusionMamba \cite{xie2024fusionmamba} \newline Apr 2024, arXiv & Image Fusion & Dynamic feature enhancement and cross-modality fusion with Mamba & Multimodal Medical and Biomedical Images & \href{https://github.com/millieXie/FusionMamba}{\faExternalLink} \\ \midrule

ReMamber \cite{yang2024remamber} \newline Mar 2024, arXiv & Referring Image Segmentation & Mamba Twister block for image-text interaction and efficient Multimodal fusion & Visual-Language Data & NA \\ \midrule

TM-Mamba \cite{wang2024text} \newline Apr 2024, arXiv & Motion Grounding & Text-controlled selection and relational embeddings for spatial graph topology & Human Motion Data & NA \\ \midrule

Cobra \cite{zhao2024cobra} \newline Mar 2024, arXiv & Multimodal LLM & Linear computational complexity and effective Multimodal Mamba fusion & Multimodal learning & \href{https://github.com/h-zhao1997/cobra}{\faExternalLink} \\ \midrule

MambaDFuse \cite{li2024mambadfuse} \newline Apr 2024, arXiv & Image Fusion & Dual-phase feature fusion and enhanced Multimodal Mamba blocks & Medical and Infrared Image Fusion & NA \\ \bottomrule

\end{tabular}%
}
\end{table*}

Multimodal models handle diverse data types like images, text, audio, and video, with the key challenge being the fusion of heterogeneous data to leverage complementary information from each modality. SurvMamba \cite{chen2024survmamba} addresses this by integrating pathological images and genomic data to improve survival predictions. It employs a Hierarchical Interaction Mamba (HIM) module to capture detailed intra-modal interactions and an Interaction Fusion Mamba (IFM) module to merge these modalities, resulting in comprehensive representations. Similarly, TransMA \cite{wu2024transma} accelerates mRNA drug delivery screening by predicting ionizable lipid nanoparticle (LNPs) properties through a 3D Transformer and molecule Mamba for feature alignment.

In large language and vision models, Meteor \cite{lee2024meteor} enhances performance by embedding detailed rationales into Multimodal frameworks. CMViM \cite{yang2024cmvim} improves Alzheimer’s disease classification by reconstructing 3D medical images through a masked Vim autoencoder and incorporating clinical data for more personalized diagnostics. SpikeMba \cite{li2024spikemba} tackles temporal video grounding using Spiking Neural Networks (SNNs), and improves both localization and contextual understanding. 

Mamba models showcase versatility in advanced applications across multiple domains. For instance, Broad Mamba enhances emotion recognition in multimodal conversations by analyzing text, video, and audio inputs \cite{shou2024revisiting}. Mamba-FETrack improves object tracking by integrating RGB video with asynchronous event streams \cite{huang2024mamba}. MambaTalk enhances gesture synthesis using video, sensor data, and audio \cite{xu2024mambatalk}, while VL-Mamba optimizes visual question answering and multimodal reasoning tasks \cite{qiao2024vl}. MambaMorph focuses on aligning MRI and CT images for precise medical analysis \cite{wang2024vmambamorph}.

For multimodal image fusion, FusionMamba \cite{xie2024fusionmamba} combines information from modalities like CT, MRI, and infrared-visible images using a Dynamic VSS block to capture global and local details. MambaDFuse \cite{li2024mambadfuse} refines this with a dual-phase feature fusion approach, improving object detection accuracy. ReMamber \cite{yang2024remamber}, TM-Mamba \cite{wang2024text}, and Cobra \cite{zhao2024cobra} integrate visual and textual data for tasks such as image segmentation and motion-text alignment. Sigma \cite{wan2024sigma} enhances semantic segmentation by fusing RGB with thermal or depth data, improving precision. Table \ref{tab:multi_modals} summarizes these models.
\section{Comparative Analysis with Traditional Frameworks}
\label{sec:comparative_analysisV}

This section provides an extensive comparison of Mamba (\textbf{\textit{M}}), CNN (\textbf{\textit{C}}), and Transformer \textbf{\textit{(T)}} models. In our analysis, we focus on key metrics such as the number of parameters (in millions, M), Floating Point Operations (FLOPs in gigaflops, G), Top-1 Accuracy (\%), Mean Intersection over Union (mIoU), Average Precision at a specific IoU threshold ($AP_x$), scalability, and performance in core computer vision tasks, including Image Classification, Object Detection, Semantic Segmentation, Video Action Classification, and Remote Sensing. The models are categorized by size—Tiny (T), Small (S), Base (B), Medium (M), Large (L), and Huge (H)—to ensure a fair comparison across different scales. By evaluating, we highlight their strengths and weaknesses to give insights about their suitability for these tasks. Before diving into the task-specific comparisons, we outline the fundamental distinctions among CNN, Transformer, and Mamba frameworks, as summarized in Table \ref{tab:cnn_transformer_mamba_fundamental_comparison}.

\begin{table*}[]
\centering
\caption{Fundamental comparison of CNN, Transformer, and Mamba Models}
\label{tab:cnn_transformer_mamba_fundamental_comparison}
\resizebox{0.85\textwidth}{!}{%
\begin{tabular}{@{}>{\centering\arraybackslash}m{3cm}>{\centering\arraybackslash}m{6.7cm}>{\centering\arraybackslash}m{6.7cm}>{\centering\arraybackslash}m{6.7cm}@{}}
\toprule
\textbf{Aspect} & \textbf{CNN} & \textbf{Transformer} & \textbf{Mamba} \\ \midrule

Core Mechanism & Convolutional layers with learnable filters \cite{krizhevsky2012imagenet} & Self-attention mechanisms \cite{vaswani2017attention} & SSMs with selective scanning operations \cite{gu2023mamba} \\  \midrule

Feature Extraction & Local feature extraction through hierarchical convolutions \cite{krizhevsky2012imagenet} & Global feature extraction through multi-head attention \cite{dosovitskiy2020image} & Combines local and global feature extraction through SSMs and dynamic scanning \cite{gu2023mamba, mamba_scanning} \\  \midrule

Receptive Field & Limited by kernel size, increases with depth \cite{krizhevsky2012imagenet,liu2024vmamba} & Global from the start due to self-attention \cite{dosovitskiy2020image,liu2024vmamba} & Global, with dynamic receptive field managed by SSMs \cite{liu2024vmamba} \\  \midrule

Positional Information & Implicitly encoded through convolutions \cite{krizhevsky2012imagenet} & Explicitly added via positional embeddings \cite{vaswani2017attention} & Implicitly modeled through SSM structure \cite{gu2023mamba, mamba_scanning} \\  \midrule

Computational Complexity & $O(kn)$ for image size $n$, where $k$ is kernel size \cite{krizhevsky2012imagenet} & $O(n^2)$ due to self-attention \cite{vaswani2017attention} & $O(n)$ with efficient SSMs and selective scanning \cite{gu2023mamba} \\ \midrule

Memory Usage & $O(n)$ for image size $n$ \cite{krizhevsky2012imagenet} & $O(n^2)$ for attention maps \cite{dosovitskiy2020image} & $O(n)$ with efficient implementation \cite{zhu2024vision} \\ \midrule

Scalability to High Resolutions & Scales well, but may lose global context \cite{liu2024vmamba} & Challenging due to quadratic complexity \cite{liu2024vmamba} & Scales efficiently due to linear complexity \cite{liu2024vmamba,zhu2024vision} \\  \midrule

Inductive Bias & Strong spatial bias \cite{krizhevsky2012imagenet} & Minimal, relies more on data \cite{dosovitskiy2020image} & Moderate, combining spatial and sequence-level patterns \cite{chaudhuri2024simba} \\  \midrule

Performance on Limited Data & Generally effective due to inductive bias and hierarchical learning \cite{li2021survey} & May struggle without sufficient data \cite{khan2022transformers} & Comparable to CNNs, but more research required \cite{gu2023mamba} \\  \midrule

Performance on Large Data & Good, but may plateau as network depth increases \cite{li2021survey} & Excellent with sufficient data \cite{khan2022transformers} & Promising, especially for long sequences \cite{liu2024vmamba} \\  \midrule

Flexibility in Feature Detection & Limited by convolutional structure \cite{krizhevsky2012imagenet} & Highly flexible due to attention mechanism \cite{vaswani2017attention} & Flexible, adaptive to various
scanning and selective mechanisms \cite{gu2023mamba} \\  \midrule

Maturity in CV Applications & Well-established \cite{li2021survey} & Increasingly adopted \cite{khan2022transformers} & Emerging, limited research so far \cite{gu2023mamba} \\  \midrule

Pre-trained Models & Abundant \cite{li2021survey} & Growing rapidly \cite{khan2022transformers} & Limited, but increasing \cite{liu2024vmamba} \\ \midrule

Training Time & Generally faster due to local processing \cite{li2021survey} & Slow for large inputs due to quadratic attention cost \cite{khan2022transformers} & Faster than Transformers for long sequences \cite{gu2023mamba} \\ \midrule

Inference Time & Generally faster\cite{li2021survey} & Typically have longer inference time \cite{khan2022transformers} & Fast, often fall between CNNs and ViTs \cite{gu2023mamba,zhu2024vision} \\ \midrule

Energy Consumption & Generally low for simple models, increases with depth \cite{krizhevsky2012imagenet} & High due to attention and large model sizes \cite{dosovitskiy2020image} & More energy-efficient due to linear time complexity\cite{gu2023mamba} \\ \midrule

\end{tabular}%
}
\end{table*}

\subsection{Image Classification} We have compared the top-performing five models from CNN, Transformer, and Mamba for image classification on ImageNet-1K dataset \cite{ILSVRC15}. The results, in Table \ref{tab:big_table_comparison_image_net_1k}, rank models based on Top-1 Accuracy (\%). For fairness, we have excluded models with over $1$ billion parameters, such as Coca \cite{yu2022coca}, and those models using extra data in the pre-training stage.
In Figure \ref{fig:all_performance_model_size}(a), we plot the Top-1 Accuracy (\%) against the number of parameters and FLOPs. The highest performers are hybrid models, while SwinV2-B \cite{hatamizadeh2024mambavision}, a Transformer model, securing $3^{rd}$ position. The top performing hybrid model Heracles-C-L \cite{patro2024heracles} achieves $1.3$ points higher Top-1 Accuracy than the top performing Transformer model SwinV2-B \cite{hatamizadeh2024mambavision}, while having $11.26$\% less FLOPS and $38.45$\% less parameters than SwinV2-B \cite{hatamizadeh2024mambavision}. This shows the dominance of hybrid Mamba models over Transformer-only models.

\begin{table*}[]
\centering

\caption{The top five models on ImageNet-1K \cite{ILSVRC15}, ranked by descending Top-1 Accuracy (\%).}

\label{tab:big_table_comparison_image_net_1k}
\resizebox{0.7\textwidth}{!}{%
\begin{tabular}{@{}>{\centering\arraybackslash}m{4cm}>{\centering\arraybackslash}m{1.4cm}>{\centering\arraybackslash}m{2cm}>{\centering\arraybackslash}m{2cm}>{\centering\arraybackslash}m{2cm}>{\centering\arraybackslash}m{2cm}>{\centering\arraybackslash}m{1cm}@{}}
\toprule
\textbf{Models} & \textbf{Backbone} & \textbf{Image Size} & \textbf{Params} & \textbf{FLOPs} & \textbf{Top-1 Acc} & \textbf{Codes} \\ \midrule
Heracles-C-L~\cite{patro2024heracles} & \textbf{\textit{T+M}} & $224^2$ & 54.1 M & 13.4 G & \textbf{85.9\% }& \href{https://github.com/badripatro/heracles}{\faExternalLink} \\ 
Heracles-C-B~\cite{patro2024heracles} & \textbf{\textit{T+M}} & $224^2$ & 32.5 M & 6.5 G & 85.2\% & \href{https://github.com/badripatro/heracles}{\faExternalLink} \\ 
SwinV2-B~\cite{liu2022swin,hatamizadeh2024mambavision} & \textbf{\textit{T}} & $256^2$ & 87.9 M & 15.1 G & 84.6\% & \href{https://github.com/microsoft/Swin-Transformer}{\faExternalLink} \\ 
Heracles-C-S~\cite{patro2024heracles} & \textbf{\textit{T+M}} & $224^2$ & \textbf{21.7 M} & \textbf{4.1 G} & 84.5\% & \href{https://github.com/badripatro/heracles}{\faExternalLink} \\ 
SiMBA-L(EinFFT)~\cite{patro2024simba} & \textbf{\textit{T+M}} & $224^2$ & 36.6 M & 7.6 G & 84.4\% & \href{https://github.com/badripatro/Simba}{\faExternalLink} \\ 
\bottomrule
\end{tabular}%
}
\end{table*}

\subsection{Object Detection}
\begin{figure*}[!t]
  \centering
  \includegraphics[scale=0.22]{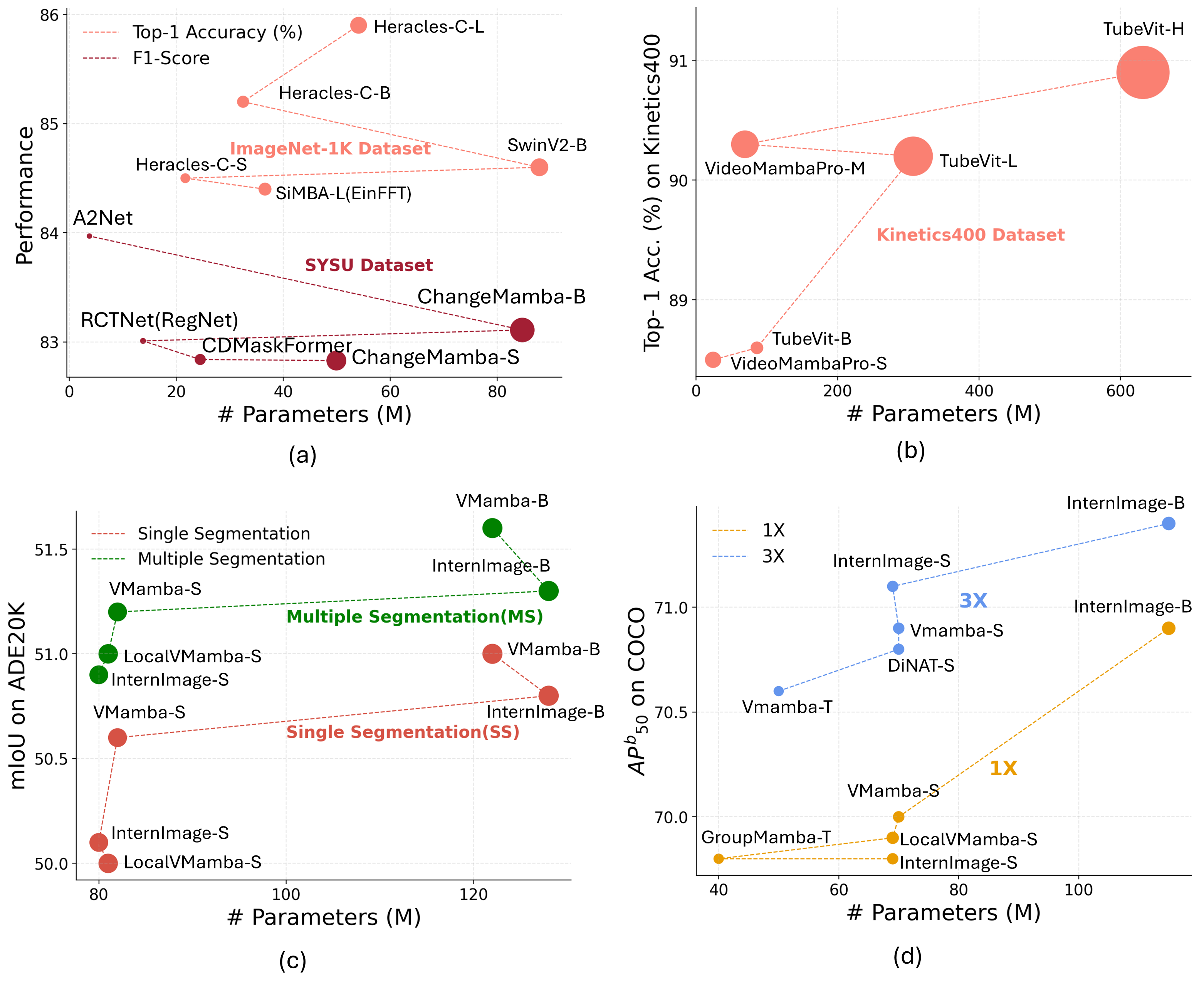}
  \caption{(a) Performance metrics for the top five models for ImageNet-1K \cite{ILSVRC15}, and SYSU dataset \cite{shi2021deeply}; (b) performance metrics against the \# parameters for Kinetics-400 dataset \cite{kay2017kinetics}; (c) performance metrics (i.e., mIoU(SS) and mIoU(MS)) against the \# parameters for ADE20K dataset \cite{zhou2017scene}; and (d) performance metrics (i.e., ${AP^{b}}_{50}$) against the \# parameters for COCO dataset \cite{lin2014microsoft}. The size of the circles represents \# FLOPs.}
  \label{fig:all_performance_model_size}
\end{figure*}

We have evaluated the top-performing five models on object detection using COCO dataset \cite{lin2014microsoft} with Mask R-CNN \cite{iccv17/mask_rcnn} framework, under $1\times$ (~12 epochs) and $3\times$ (~36 epochs) training schedules. We have ranked the models based on ${AP^{b}}_{50}$ scores. The results, summarized in Table \ref{tab:big_table_comparison_ms_coco} and Figure \ref{fig:all_performance_model_size}(d).
From Figure \ref{fig:all_performance_model_size}(d), it is evident that the Mamba models, VMamba-S \cite{liu2024vmamba}, LocalVMamba-S \cite{huang2024localmamba}, and GroupMamba-T \cite{shaker2024groupmamba} are among the top performers for the 1$\times$schedule. Notably, GroupMamba-T \cite{shaker2024groupmamba} achieves just $1.1$ points lower ${AP^{b}}_{50}$ than the best model, InternImage-B \cite{wang2023internimage}, while using $65.21$\% fewer parameters and $44.31$\% less FLOPS.
For the $3\times$schedule, VMamba-S \cite{liu2024vmamba} and VMamba-T \cite{liu2024vmamba} are strong competitors among the top-performing five models. Specifically, VMamba-T achieves only $0.8$ points lower ${AP^{b}}_{50}$ compared to InternImage-B \cite{wang2023internimage} while consuming $56.52$\% fewer parameters and $45.91$\% less FLOPS. This demonstrates that the VMamba maintains competitive performance with significantly lower computational costs, making it suitable for resource-constrained environments.

\begin{table*}[]
\centering
\caption{The top five models on COCO \textit{minival} dataset \cite{lin2014microsoft} for object detection and segmentation using Mask R-CNN \cite{iccv17/mask_rcnn} with $1\times$ and $3\times$ schedules. FLOPs are calculated with input size $1280\times800$. $AP^{b}$ and $AP^{m}$ represent box AP and mask AP, respectively, and the results are ranked by descending $AP^{b}_{50}$.}

\label{tab:big_table_comparison_ms_coco}
\resizebox{0.65\textwidth}{!}{%
\begin{tabular}{@{}>{\centering\arraybackslash}m{3.1cm}>{\centering\arraybackslash}m{1.4cm}>{\centering\arraybackslash}m{1cm}>{\centering\arraybackslash}m{1cm}>{\centering\arraybackslash}m{.8cm}>{\centering\arraybackslash}m{.8cm}>{\centering\arraybackslash}m{.8cm}>{\centering\arraybackslash}m{.8cm}>{\centering\arraybackslash}m{.8cm}>{\centering\arraybackslash}m{.8cm}>{\centering\arraybackslash}m{1cm}@{}}
\toprule

\textbf{Models} & \textbf{Backbone} & \textbf{Params} & \textbf{FLOPs} & \textbf{$AP^b$} & \textbf{$AP^{b}_{50}$} & \textbf{$AP^{b}_{75}$} & \textbf{$AP^m$} & \textbf{$AP^{m}_{50}$} & \textbf{$AP^{m}_{75}$} & \textbf{Codes} \\ \midrule

\multicolumn{11}{c}{\cellcolor{lightgray}{\textbf{Mask R-CNN 1$\times$schedule}}} \\ \midrule

InternImage-B~\cite{wang2023internimage} & \textbf{\textit{C}} & 115 M & 501 G & \textbf{48.8} & \textbf{70.9} & \textbf{54.0} & \textbf{44.0} & \textbf{67.8} & \textbf{47.4} & \href{https://github.com/OpenGVLab/InternImage}{\faExternalLink} \\ 
VMamba-S~\cite{liu2024vmamba} & \textbf{\textit{M}} & 70 M & 349 G & 48.7 & 70.0 & 53.4 & 43.7 & 67.3 & 47.0 & \href{https://github.com/MzeroMiko/VMamba}{\faExternalLink} \\ 
LocalVMamba-S~\cite{huang2024localmamba} & \textbf{\textit{M}} & 69 M & 414 G & 48.4 & 69.9 & 52.7 & 43.2 & 66.7 & 46.5 & \href{https://github.com/hunto/LocalMamba}{\faExternalLink} \\ 
GroupMamba-T~\cite{shaker2024groupmamba} & \textbf{\textit{M}} & \textbf{40 M} & 2\textbf{79 G} & 47.6 & 69.8 & 52.1 & 42.9 & 66.5 & 46.3 & \href{https://github.com/Amshaker/GroupMamba}{\faExternalLink} \\ 
InternImage-S~\cite{wang2023internimage} & \textbf{\textit{C}} & 69 M & 340 G & 47.8 & 69.8 & 52.8 & 43.3 & 67.1 & 46.7 & \href{https://github.com/OpenGVLab/InternImage}{\faExternalLink} \\ \midrule

\multicolumn{11}{c}{\cellcolor{lightgray}{\textbf{Mask R-CNN 3$\times$schedule}}} \\ \midrule

InternImage-B~\cite{wang2023internimage} & \textbf{\textit{C}} & 115 M & 501 G & \textbf{50.3} & \textbf{71.4} & \textbf{55.3} & \textbf{44.8} & \textbf{68.7} & \textbf{48.0} & \href{https://github.com/OpenGVLab/InternImage}{\faExternalLink} \\ 
InternImage-S~\cite{wang2023internimage} & \textbf{\textit{C}} & 69 M & 340 G & 49.7 & 71.1 & 54.5 & 44.5 & 68.5 & 47.8 & \href{https://github.com/OpenGVLab/InternImage}{\faExternalLink} \\ 
VMamba-S~\cite{liu2024vmamba} & \textbf{\textit{M}} & 70 M & 349 G & 49.9 & 70.9 & 54.7 & 44.2 & 68.2 & 47.7 & \href{https://github.com/MzeroMiko/VMamba}{\faExternalLink} \\ 
DiNAT-S~\cite{hassani2022dilated} & \textbf{\textit{T}} & 70 M & 330 G & 49.3 & 70.8 & 54.2 & 44.0 & 68.0 & 47.4 & \href{https://github.com/SHI-Labs/Neighborhood-Attention-Transformer}{\faExternalLink} \\ 
VMamba-T~\cite{liu2024vmamba} & \textbf{\textit{M}} & \textbf{50 M} & 2\textbf{71 G} & 48.8 & 70.6 & 53.6 & 43.7 & 67.7 & 46.8 & \href{https://github.com/MzeroMiko/VMamba}{\faExternalLink} \\ \bottomrule
\end{tabular}%
}
\end{table*}

\subsection{Semantic Segmentation}
We have analyzed the top-performing five models for semantic segmentation. The results are shown in Table \ref{tab:big_table_comparison_ade_20k} and Figure \ref{fig:all_performance_model_size}(c). They present the mIoU scores for both single segmentation (SS) and multiple segmentation (MS) settings. In our analysis, we have excluded any models that used additional datasets for pre-training to ensure fairness.
In Table \ref{tab:big_table_comparison_ade_20k}, the Mamba model VMamba-B \cite{liu2024vmamba} outperforms the previous CNN-based SOTA model InternImage-B \cite{wang2023internimage} by $0.2$ points in mIoU (SS) and $0.3$ points in mIoU (MS) with similar computational costs.
The VMamba-S \cite{liu2024vmamba} and LocalVMamba-S \cite{huang2024localmamba} models also show competitive mIoU scores with relatively lower parameter counts and FLOPs. Notably, VMamba-S achieves a mIoU (SS) of $50.6$ and mIoU (MS) of $51.2$, while LocalVMamba-S scores $50.0$ and $51.0$, respectively. 
Interestingly, none of the Transformer-based models make it into the top-performing five comparisons, indicating that the CNN and Mamba models could be an effective choice for semantic segmentation tasks.

\begin{table*}[!t]
\centering
\caption{The top five models on ADE20K $val$ dataset \cite{zhou2017scene} are ranked by mIoU (SS) for semantic segmentation using the UperNet framework \cite{eccv18/upernet}. A crop size of $512^2$ is used, and FLOPs are calculated with an input size of $512\times2048$. Models using extra pre-training data are excluded for fairness.}

\label{tab:big_table_comparison_ade_20k}
\resizebox{0.7\textwidth}{!}{%
\begin{tabular}{@{}>{\centering\arraybackslash}m{4cm}>{\centering\arraybackslash}m{1.4cm}>{\centering\arraybackslash}m{2cm}>{\centering\arraybackslash}m{2cm}>{\centering\arraybackslash}m{2cm}>{\centering\arraybackslash}m{2cm}>{\centering\arraybackslash}m{1cm}@{}}
\toprule
\textbf{Models} & \textbf{Backbone} & \textbf{Params} & \textbf{FLOPs} & \textbf{mIoU (SS)} & \textbf{mIoU (MS)} & \textbf{Codes} \\ \midrule
VMamba-B~\cite{liu2024vmamba} & \textbf{\textit{M}} & 122 M & 1170 G & \textbf{51.0} & \textbf{51.6} & \href{https://github.com/MzeroMiko/VMamba}{\faExternalLink} \\ 
InternImage-B~\cite{wang2023internimage} & \textbf{\textit{C}} & 128 M & 1185 G & 50.8 & 51.3 & \href{https://github.com/OpenGVLab/InternImage}{\faExternalLink} \\ 
VMamba-S~\cite{liu2024vmamba} & \textbf{\textit{M}} & 82 M & 1028 G & 50.6 & 51.2 & \href{https://github.com/MzeroMiko/VMamba}{\faExternalLink} \\ 
InternImage-S~\cite{wang2023internimage} & \textbf{\textit{C}} & \textbf{80 M} & \textbf{1017 G} & 50.1 & 50.9 & \href{https://github.com/OpenGVLab/InternImage}{\faExternalLink} \\ 
LocalVMamba-S~\cite{huang2024localmamba} & \textbf{\textit{M}} & 81 M & 1095 G & 50.0 & 51.0 & \href{https://github.com/hunto/LocalMamba}{\faExternalLink} \\ \bottomrule
\end{tabular}%
}
\end{table*}

\subsection{Video Action Classification}
We have considered the Kinetics 400 dataset \cite{kay2017kinetics} as a popular dataset for video action classification with 400 human action classes. We present the top-performing five models based on Top-1 accuracy (\%) in Table \ref{tab:big_table_comparison_kinetics_400} and Figure \ref{fig:all_performance_model_size}(b) along with their parameter and FLOP sizes.
The Mamba model VideoMambaPro-M \cite{lu2024videomambapro} is the second best model which is $0.6$ and $0.4$ points behind the Transformer-based top model TubeVit-H \cite{piergiovanni2023rethinking} in terms of Top-1 and Top-5 accuracies (\%), respectively. However, it consumes $89.08$\% fewer parameters and $73.92$\% fewer FLOPs than TubeVit-H. Therefore, it demonstrates the superiority in computational efficiency for Mamba-based models over Transformer-based models.
Other models like TubeVit-L and TubeVit-B \cite{piergiovanni2023rethinking} also perform well with higher computational costs. VideoMambaPro-S \cite{lu2024videomambapro} achieves a Top-1 accuracy of $88.5$\% with low parameter counts and FLOPs, further showcasing Mamba's efficiency.

\begin{table*}[]
\centering
\caption{The top five models for video action classification on Kinetics-400 dataset \cite{kay2017kinetics} are ranked by Top-1 Accuracy (\%) under supervised training, with only ImageNet-1K (IN-1K) \cite{ILSVRC15} pre-trained models considered.}
\label{tab:big_table_comparison_kinetics_400}
\resizebox{0.7\textwidth}{!}{%
\begin{tabular}{@{}>{\centering\arraybackslash}m{3.8cm}>{\centering\arraybackslash}m{1.4cm}>{\centering\arraybackslash}m{1.5cm}>{\centering\arraybackslash}m{2cm}>{\centering\arraybackslash}m{1cm}>{\centering\arraybackslash}m{1.2cm}>{\centering\arraybackslash}m{1cm}>{\centering\arraybackslash}m{1cm}>{\centering\arraybackslash}m{1cm}@{}}

\toprule
\textbf{Models} & \textbf{Backbone} & \textbf{Extra Data} & \textbf{Input Size} & \textbf{Params} & \textbf{FLOPs} & \textbf{Top-1 Acc} & \textbf{Top-5 Acc} & \textbf{Codes} \\ \midrule

TubeVit-H~\cite{piergiovanni2023rethinking} & \textbf{\textit{T}} & IN-1K & $32\times224^2$ & 632 M & 17640 G &\textbf{ 90.9\%} & \textbf{98.9\% }& \href{https://github.com/daniel-code/TubeViT}{\faExternalLink} \\ 
VideoMambaPro-M~\cite{lu2024videomambapro} & \textbf{\textit{M}} & IN-1K & $32\times224^2$ & 69 M & 4600 G & 90.3\% & 98.5\% & \href{https://github.com/hotfinda/VideoMambaPro}{\faExternalLink} \\ 
TubeVit-L~\cite{piergiovanni2023rethinking} & \textbf{\textit{T}} & IN-1K & $32\times224^2$ & 307 M & 9530 G & 90.2\% & 98.6\% & \href{https://github.com/daniel-code/TubeViT}{\faExternalLink} \\ 
TubeVit-B~\cite{piergiovanni2023rethinking} & \textbf{\textit{T}} & IN-1K & $32\times224^2$ & 86 M & \textbf{870 G} & 88.6\% & 97.6\% & \href{https://github.com/daniel-code/TubeViT}{\faExternalLink} \\ 
VideoMambaPro-S~\cite{lu2024videomambapro} & \textbf{\textit{M}} & IN-1K & $32\times224^2$ & \textbf{24 M} & 1500 G & 88.5\% & 98.1\% & \href{https://github.com/hotfinda/VideoMambaPro}{\faExternalLink} \\ \bottomrule
\end{tabular}%
}
\end{table*}

\subsection{Remote Sensing}
We have evaluated the top-performing five models based on F1 Scores for binary change detection in remote sensing using the SYSU-CD dataset \cite{shi2021deeply}. The results are summarized in Table \ref{tab:big_table_comparison_sysu} and Figure \ref{fig:all_performance_model_size}(a).
ChangeMamba-B \cite{chen2024changemamba} ranks as the second-best model with an F1 score $0.86$ points lower than the Transformer-based SOTA, A2Net \cite{li2023lightweight}. However, ChangeMamba-B has significantly higher parameter counts and FLOPs, indicating that Mamba models currently lack computational efficiency for this task.
Additionally, combining CNN and Transformer, RCTNet (RegNet) \cite{gao2024relating} and CDMaskFormer \cite{ma2024rethinking} also perform well but show more balance in terms of computational costs compared to ChangeMamba models. ChangeMamba-S \cite{chen2024changemamba}, another Mamba variant, achieves competitive F1 scores but still consumes substantially more resources than other top models. 

\begin{table*}[!h]
\centering
\caption{The top five models on binary remote sensing change detection for SYSU \cite{shi2021deeply} dataset based on \textbf{F1} scores. The results are shown in decreasing order of F1 scores. Here, \textbf{R} = Recall, \textbf{P} = Precision.}
\label{tab:big_table_comparison_sysu}
\resizebox{0.7\textwidth}{!}{%
\begin{tabular}{@{}>{\centering\arraybackslash}m{3.8cm}>{\centering\arraybackslash}m{1.4cm}>{\centering\arraybackslash}m{1.5cm}>{\centering\arraybackslash}m{2cm}>{\centering\arraybackslash}m{1cm}>{\centering\arraybackslash}m{1cm}>{\centering\arraybackslash}m{1cm}>{\centering\arraybackslash}m{1cm}>{\centering\arraybackslash}m{1cm}@{}}

\toprule
\textbf{Model} & \textbf{Backbone} & \textbf{Params} & \textbf{FLOPs} & \textbf{R} & \textbf{P} & \textbf{F1} & \textbf{IoU} & \textbf{Code} \\ \midrule

A2Net~\cite{li2023lightweight} & \textbf{\textit{C}} & \textbf{3.78 M} & \textbf{6.02 G} & 82.24 & 85.77 & \textbf{83.97} & \textbf{72.37} & \href{https://github.com/guanyuezhen/A2Net}{\faExternalLink} \\ 
ChangeMamba-B~\cite{chen2024changemamba} & \textbf{\textit{M}} & 84.70 M & 179.32 G & 80.31 & 86.11 & 83.11 & 71.10 & \href{https://github.com/ChenHongruixuan/MambaCD}{\faExternalLink} \\ 
RCTNet (RegNet)~\cite{gao2024relating} & \textbf{\textit{C+T}} & 13.8 M & 7.65 G & 81.73 & 84.33 & 83.01 & 70.96 & \href{https://github.com/NUST-Machine-Intelligence-Laboratory/RCTNet}{\faExternalLink} \\ 
CDMaskFormer~\cite{ma2024rethinking} & \textbf{\textit{C+T}} & 24.49 M & 32.46 G & \textbf{87.25} & 78.85 & 82.84 & 70.70 & \href{https://github.com/xwmaxwma/rschange}{\faExternalLink} \\ 
ChangeMamba-S~\cite{chen2024changemamba} & \textbf{\textit{M}} & 49.94 M & 114.82 G & 78.25 & \textbf{87.99} & 82.83 & 70.70 & \href{https://github.com/ChenHongruixuan/MambaCD}{\faExternalLink} \\ \bottomrule
\end{tabular}%
}
\end{table*}

\section{Potential Limitations and Future Prospects}
\label{sec:limitations_mamba_V}

Despite the significant advancements and promising capabilities of Mamba models, several limitations hinder their broader adaption and optimal performance. This section outlines these potential limitations and future prospects to ensure Mamba models can achieve their full potential.

\subsection{Limited Generalizability: Domain-Specific Biases and Hidden State Accumulation}

Despite Mamba's global receptive field, it often struggles to generalize across different domains. This limitation arises from two key factors. First, the selective scanning process captures domain-specific information in hidden states \cite{long2024dgmamba}, creating compressed representations biased toward the training data, which limits the model’s adaptability to new domains. Second, scanning methods such as bi-directional scanning often reinforce domain-specific biases \cite{zhu2024vision}. For example, models trained on natural images may prioritize texture patterns, while those trained on medical scans may focus on anatomical shapes, limiting to learn domain-agnostic features \cite{huix2024natural}.

To address hidden state accumulation, dropout layers or weight normalization could be applied directly to hidden states within the Mamba architecture. These techniques introduce controlled noise or constraints during training to help the model learn more generalizable representations \cite{gu2023mamba}. Additionally, developing new scanning mechanisms that avoid capturing domain-specific biases is crucial. Approaches like domain-adaptive scanning, which adjusts based on input, or incorporating learnable masks to focus on relevant features selectively, could improve generalizability \cite{ling2023domain}.

\subsection{Challenges in Selecting an Effective Scanning Mechanism}

Mamba, originally designed for 1D sequential data, facees significant challenges when adapting their selective scanning methods to multi-dimensional visual data \cite{huang2024localmamba}. One major challenge is accurately capturing images' complex spatial dependencies and hierarchical structures. Traditional scanning techniques, such as linear or raster scanning \cite{zhu2024vision}, often fail to preserve the intricate spatial relationships needed for detailed image analysis. This limitation arises from the mismatch between Mamba's 1D sequential processing and the inherently multi-dimensional nature of visual information. Moreover, the redundancy of scanning in multiple directions increases computational demands, further complicating the adaptation to visual data.

To improve Mamba's effectiveness in vision, several promising strategies have been explored \cite{zhu2024vision, hu2024zigma, yang2024vivim, pei2024efficientvmamba}. Developing multi-dimensional selective SSMs could make visual data processing more efficient while preserving Mamba's computational benefits. Introducing hierarchical scanning patterns inspired by human visual processing or using attention-guided scanning mechanisms could enhance the model's ability to handle complex visual data \cite{zhang2024motion}.
Additionally, sparse scanning techniques that focus on processing only the most informative patches of an image. It optimizes computational efficiency and performance for large-scale vision tasks \cite{pei2024efficientvmamba}. An experimental study on different scanning techniques provides valuable insights into these strategies \cite{mamba_scanning}.

\subsection{Limited Pre-trained Model Availability and Community Support}
The adaptation of deep learning architectures heavily rely on the availability of pre-trained models. Currently, Mamba has a limited selection of pre-trained models compared to more established architectures like Transformers. Although a notable $2.8$B-parameter Mamba model exists \cite{gu2023mamba}, the overall variety and number of pre-trained models are still limited. This scarcity limits their applicability for various downstream tasks. Additionally, the Mamba research community is relatively new, with fewer researchers actively contributing to its development. This situation slows the pace of innovation and limits available resources for developers.

To expand the availability of pre-trained models, large-scale pre-training on diverse datasets is required. These pre-trained models enable fine-tuning for specific tasks to reduce training times and improve overall performance. Additionally, fostering collaboration and knowledge sharing within the Mamba research community is crucial. Initiatives like organizing workshops, developing open-source repositories, and creating forums for discussion can foster growth, enhance knowledge exchange, and accelerate the architecture’s development \cite{haven2023mambachat}.

\subsection{Interpretability and Explainability: Unveiling the Black Box}
Mamba demonstrates impressive performance on various computer vision tasks, however it struggles with interpretability and explainability. The complex sequential nature of Mamba's SSM, coupled with non-linear activations and selective state updates, complicates tracing the model's decision-making process or identifying which features most influence its predictions \cite{ali2024hidden}. The selective scanning mechanism in Mamba is inherently non-linear, making it challenging to pinpoint the exact sequence of computations that lead to a specific prediction. 
Moreover, the hidden states capture complex, compressed representations of processed data, making it challenging to interpret the specific features and relationships encoded within these high-dimensional spaces \cite{xing2024segmamba}.

Recent research \cite{ali2024hidden, han2024demystify} has made progress in unveiling Mamba's internal processes. One study reinterpreted Mamba layers as implicit self-attention mechanisms, revealing hidden attention matrices within the model \cite{ali2024hidden}. This finding allows the adaptation of explainability techniques developed for Transformers to Mamba architectures.
Researchers are exploring various methods to enhance interpretability, such as attention-inspired visualization techniques \cite{rahman2020efficient} to highlight important spatial and temporal patterns in the input data, feature attribution methods to identify the most influential input features, and state analysis by intervening on SSM states to understand information flow \cite{han2024demystify}. Additionally, employing model-agnostic explainability methods can offer further insights into Mamba's decision-making process \cite{selvaraju2017grad}.

\subsection{Security and Adversarial Robustness}
VMamba \cite{liu2024vmamba} faces notable security challenges, particularly its vulnerability to adversarial attacks. Recent study\cite{du2024understanding} has demonstrated that VMamba is susceptible to both whole-image and patch-specific adversarial manipulations, which can alter model predictions. However, VMamba exhibits stronger adversarial robustness than Transformer architectures, especially in smaller models \cite{du2024understanding}.

Prior work \cite{du2024understanding} has identified the parameters $B$ and $C$ in the SSM blocks as key vulnerability points, while the $\Delta$ parameter provides some defense against white-box attacks. To enhance robustness, future work should focus on reducing reliance on $B$ and $C$, strengthening the defensive role of $\Delta$, and developing adversarial training techniques tailored to VMamba. Moreover, addressing VMamba's scalability issues—since its robustness diminishes with increasing model complexity—is critical for ensuring the security of larger-scale applications \cite{du2024understanding, malik2024towards}.
\section{Conclusion}
\label{sec:conclusion_V}

In this work, we provide a comprehensive overview of recent applications of Mamba in computer vision. We begin by discussing the limitations of traditional architectures, particularly CNN and Transformer. Most importantly, we highlight the trade-offs between Mamba and traditional architectures, addressing constraints such as quadratic complexities, inductive bias, and long-range dependencies. Moreover, we offer a structured taxonomy of Mamba applications, present a generalized pipeline, and visualize the various scanning methods used within Mamba models. Besides, we present the specific strengths, weaknesses, and potential use cases of different scanning methods. Then we conduct a comparative analysis across different application areas, with quantitative evaluations using various datasets, demonstrating the performance of Mamba models compared to traditional architectures. Finally, we identify a set of critical challenges to inspire research and further advance this emerging field. We hope this survey serves as a valuable reference for Mamba and offers insights that inspire innovation and guide future developments for researchers.

\bibliographystyle{ACM-Reference-Format}
\bibliography{bibliography}

\end{document}